\documentclass{article}

\usepackage[final]{neurips_2022}

\usepackage[colorlinks]{hyperref}
\usepackage{url}
\usepackage{amsfonts}
\usepackage{nicefrac}
\usepackage{microtype}
\usepackage{xcolor} 
\usepackage{amsmath}
\usepackage{amssymb}
\usepackage{mathtools}
\usepackage{booktabs}
\usepackage{ragged2e}
\usepackage{tikz}
\usepackage{stackengine}
\usepackage{etoolbox}
\usepackage{xspace}
\usepackage{xpatch}
\usepackage{cuted}
\usepackage{enumerate}
\usepackage{xstring}
\usepackage{natbib}
\usepackage{setspace}
\usepackage{tabularx}
\usepackage{makecell}
\usepackage{graphicx}
\usepackage{changepage}
\usepackage{enumitem}
\usepackage{cuted}
\usepackage{titlesec}
\usepackage{cancel}
\usepackage{eqparbox}
\usepackage{wrapfig}
\usepackage[hang,flushmargin,bottom]{footmisc}
\usepackage[capitalise,noabbrev,nameinlink]{cleveref}
\usepackage[useregional=numeric]{datetime2}
\usepackage{tocbasic}
\usepackage{eqparbox}
\usepackage{pifont}
\usepackage{listings}
\usepackage{todonotes}

\usetikzlibrary{positioning,arrows.meta,fit,calc,backgrounds}
\tikzset{%
node distance=2em, auto,
every node/.style={line width=0.7pt},
det/.style={draw=black, rectangle, minimum size=2.5em, inner sep=0.1ex},
lat/.style={draw=black, circle, minimum size=2.5em, inner sep=0.1ex},
obs/.style={draw=black, circle, fill=black!15, minimum size=2.5em, inner sep=0.1ex},
fac/.style={draw=black, rectangle, fill=black, minimum size=.6em, inner sep=0em},
dummy/.style={draw=none, circle, minimum size=2.5em},
plate/.style={draw=black, rounded corners, inner sep=.8em, yshift=-.7em, align=right},
box/.style={draw=black, rounded corners, inner sep=.4em, align=center},
generates/.style={->, -{Stealth[length=.6em, inset=0pt]}, line width=0.7pt},
undirected/.style={line width=0.7pt},
}

\setlist[itemize]{leftmargin=1.5em,itemsep=.1em,topsep=.1em}
\titlespacing*{\paragraph}{0pt}{0ex plus .1ex}{1em}
\DeclareTOCStyleEntry[beforeskip=.2em plus 1pt]{tocline}{section}
\xapptocmd\normalsize{%
\abovedisplayskip=.8em plus .2em minus .2em
\belowdisplayskip=.6em plus .1em minus .1em
\abovedisplayshortskip=.8em plus .2em minus .2em
\belowdisplayshortskip=.6em plus .1em minus .1em
}{}{}

\newcommand{\padspace}{\hspace{3.5em}}

\setcitestyle{authoryear,round,citesep={;},aysep={,},yysep={;}}
\renewcommand{\cite}[1]{\citep{#1}}

\creflabelformat{equation}{#2\textup{#1}#3}
\crefname{algocf}{Algorithm}{Algorithms}
\Crefname{algocf}{Algorithm}{Algorithms}

\definecolor{mydarkblue}{rgb}{0,0.08,0.45}
\hypersetup{%
colorlinks=true,
linkcolor=mydarkblue,
citecolor=mydarkblue,
filecolor=mydarkblue,
urlcolor=mydarkblue}

\graphicspath{{figures/}}

\makeatletter
\def\hlinewd#1{%
\noalign{\ifnum0=`}\fi\hrule \@height #1 \futurelet
\reserved@a\@xhline}
\makeatother

\makeatletter
\DeclareDocumentCommand\todo{g}{%
\def\@message{\IfNoValueTF{#1}{TODO}{TODO: #1}}
\textbf{\textcolor[HTML]{FF8811}{\@message}}
\@latex@warning{\@message}{}{}}
\makeatother

\newcommand{\describe}[3][0pt]{\hspace*{.12em}\underbracket[0.5pt][1pt]{#2\hspace*{#1}}_\text{#3}}

\makeatletter
\newcommand{\removeParBefore}{\ifvmode\vspace*{-\baselineskip}\setlength{\parskip}{0ex}\fi}
\newcommand{\removeParAfter}{\@ifnextchar\par\@gobble\relax}
\newcommand{\eq}{\begingroup\removeParBefore\endlinechar=32 \eqinner}
\newcommand{\eqinner}[2][aligned]{\endlinechar=32%
\begin{gather}\begin{#1}#2\end{#1}\end{gather}\endgroup\removeParAfter}
\makeatother

\DeclareDocumentCommand{\p}{ D<>{p} D<>{} r() }{
\def\content{#3}\patchcmd{\content}{|}{\;#2\vert\;}{}{}
\ensuremath{#1 #2(\content #2)}}

\DeclareDocumentCommand{\P}{ D<>{P} D<>{\big} r() }{
\def\content{#3}\patchcmd{\content}{|}{\;#2\vert\;}{}{}
\ensuremath{\operatorname{#1}#2(\content #2)}}

\DeclareDocumentCommand{\E}{ D<>{E} E{_}{{}} D<>{\big} r[] }{
\def\content{#4}\patchcmd{\content}{|}{\;#3\vert\;}{}{}
\ensuremath{\operatorname{#1}_{#2}#3[\content #3]}}

\DeclareDocumentCommand{\D}{ D<>{D} D<>{\big} r[] }{
\def\content{#3}\patchcmd{\content}{||}{\;#2\|\;}{}{}
\ensuremath{\operatorname{#1}\!#2[\content #2]}}

\NewDocumentCommand{\Nor}{ r() }{\P<Normal>](#1)}
\NewDocumentCommand{\Cat}{ r() }{\P<Cat>](#1)}
\NewDocumentCommand{\Bin}{ r() }{\P<Bin>](#1)}
\NewDocumentCommand{\Bet}{ r() }{\P<Beta>](#1)}
\NewDocumentCommand{\Ber}{ r() }{\P<Bernoulli>(#1)}
\NewDocumentCommand{\Dir}{ r() }{\P<Dir>(#1)}

\DeclareDocumentCommand{\KL}{ D<>{\big} r[] }{\D<KL><#1>[#2]}
\DeclareDocumentCommand{\H}{ D<>{\big} r[] }{\E<H><#1>[#2]}
\DeclareDocumentCommand{\I}{ D<>{\big} r[] }{\E<I><#1>[#2]}

\DeclareDocumentCommand{\pp}{ D<>{} r() }{
\ensuremath{\p<p_\phi><#1>(#2)}}
\DeclareDocumentCommand{\qp}{ D<>{} r() }{
\ensuremath{\p<q_\phi><#1>(#2)}}

\usepackage[utf8]{inputenc} 
\usepackage[T1]{fontenc}    
\usepackage{hyperref}       
\usepackage{url}            
\usepackage{booktabs}       
\usepackage{amsfonts}       
\usepackage{nicefrac}       
\usepackage{microtype}      
\usepackage{xcolor}         
\usepackage{subcaption}
\usepackage{xcolor}

\title{Learning Robust Dynamics through\\Variational Sparse Gating}

\author{%
    Arnav Kumar Jain$^{1, 2, *}$,
    Shivakanth Sujit$^{2, 3}$,
    Shruti Joshi$^{2, 3}$,
    Vincent Michalski$^{1, 2}$\\
    \textbf{Danijar Hafner}$^{4, 5}$,
    \textbf{Samira Ebrahimi-Kahou}$^{2, 3, 6}$
}

\usepackage[acronym]{glossaries}
\glsdisablehyper

\newcommand{\env}{BringBackShapes}
\newacronym{cnn}{CNN}{Convolutional Neural Network}
\newacronym{elu}{ELU}{Exponential Linear Unit}
\newacronym{gru}{GRU}{Gated Recurrent Unit}
\newacronym{lstm}{LSTM}{Long Short-term Memory}
\newacronym{mse}{MSE}{mean-squared error}
\newacronym{mlp}{MLP}{Multi-layer Perceptron}
\newacronym{rl}{RL}{Reinforcement Learning}
\newacronym{rim}{RIM}{Recurrent Independent Mechanism}
\newacronym{rnn}{RNN}{Recurrent Neural Network}
\newacronym{rssm}{RSSM}{Recurrent State-Space Model}
\newacronym{ssm}{SSM}{Stochastic State-Space Model}
\newacronym{alg}{VSG}{Variational Sparse Gating}
\newacronym{bbs}{BBS}{BringBackShapes}
\newacronym{algsimple}{SVSG}{Simple Variational Sparse Gating}

\newcommand{\alg}{\gls{alg}}
\newcommand{\algsimple}{\gls{algsimple}}
\newcommand\blfootnote[1]{%
  \begingroup
  \renewcommand\thefootnote{}\footnote{#1}%
  \addtocounter{footnote}{-1}%
  \endgroup
}

\begin{document}

\maketitle

\begin{abstract}
Learning world models from their sensory inputs enables agents to plan for actions by imagining their future outcomes. World models have previously been shown to improve sample-efficiency in simulated environments with few objects, but have not yet been applied successfully to environments with many objects. In environments with many objects, often only a small number of them are moving or interacting at the same time. In this paper, we investigate integrating this inductive bias of sparse interactions into the latent dynamics of world models trained from pixels. First, we introduce Variational Sparse Gating~(VSG), a latent dynamics model that updates its feature dimensions sparsely through stochastic binary gates. Moreover, we propose a simplified architecture Simple Variational Sparse Gating~(SVSG) that removes the deterministic pathway of previous models, resulting in a fully stochastic transition function that leverages the VSG mechanism. We evaluate the two model architectures in the BringBackShapes~(BBS) environment that features a large number of moving objects and partial observability, demonstrating clear improvements over prior models.
\end{abstract}

\blfootnote{
 $^{1}$Université de Montréal, $^{2}$Mila- Quebec AI Institute, $^{3}$École de technologie supérieure, $^{4}$University of Toronto, $^{5}$Google Brain, $^{6}$CIFAR. $^{*}$Correspondence to Arnav Kumar Jain <arnav-kumar.jain@mila.quebec>. 
 }

\section{Introduction}
\label{Sec:Introduction}
Latent dynamics models are models that generate agent's future states in the compact latent space without feeding the high-dimensional observations back to the model. They have shown promising results on various tasks like video prediction~\cite{karl2016dvbf,kalman1960filter,krishnan2015deepkalman}, model-based \gls{rl} ~\citep{hafner2020dreamer,hafner2021mastering,hafner2019planning,ha2018worldmodels}, and robotics~\cite{watter2015e2c}. Generating sequences in the compact latent space reduces the accumulating errors leading to more accurate long-term predictions. Additionally, having lower dimensionality leads to a lower memory footprint. Solving tasks in model-based \gls{rl} involves learning a world model~\citep{ha2018worldmodels} that can predict outcomes of actions, followed by using them to derive behaviors~\citep{sutton1991dyna}. 
Motivated by these benefits, the recently proposed DreamerV1~\cite{hafner2020dreamer} and DreamerV2~\cite{hafner2021mastering} agents achieved state-of-the-art results on a wide range of visual control tasks.

Many complex tasks require reliable long-term prediction of dynamics. This is true especially in partially observable environments where only a subspace is visible to the agent, and it is usually required to accurately retain information over multiple time steps to solve the task. The Dreamer agents~\citep{hafner2020dreamer, hafner2021mastering} employ an \gls{rssm}~\cite{hafner2019planning} comprising of a \gls{rnn}. Training \glspl{rnn} for long sequences is challenging as they suffer from optimization problems like vanishing gradients~\citep{hochreiter1991untersuchungen,bengio1994learning}. Different ways of applying sparse updates in \glspl{rnn} have been investigated~\cite{campos2017skip,neil2016phased,goyal2019recurrent}, enabling a subset of state dimensions to be constant during the update. A sparse update prior can also be motivated by the fact that in the real world, many factors of variation are constant over extended periods of time. 
For instance, several objects in a physical simulation may be stationary until some force acts upon them.
Additionally, this is useful in the partially observable setting where the agent observes a constrained viewpoint and has to keep track of objects that are not visible for many time steps. 
In this work, we introduce \glsfirst{alg}, a stochastic gating mechanism that sparsely updates the latent states at each step. 

\acrfull{rssm}~\citep{hafner2019planning} was introduced in PLaNet where the model state was composed of two paths, an image representation path and a recurrent path. DreamerV1~\cite{hafner2020dreamer} and DreamerV2~\cite{hafner2021mastering} utilized them to achieve state-of-the-art results in continuous and discrete control tasks~\cite{hafner2019planning}. 
While the image representation path which is stochastic accounts for multiple possible future states, the recurrent path is deterministic to retain information over multiple time steps to facilitate gradient-based optimization. 
\cite{hafner2019planning} showed that both components were important for solving tasks, where the stochastic part was more important to account for partial observability of the initial states. 
By leveraging the proposed gating mechanism (\alg), we demonstrate that a purely stochastic model with a single component can achieve competitive results, and call it \algsimple.
To the best of our knowledge, this is the first work that shows that purely stochastic models achieve competitive performance on continuous control tasks when compared to leading agents.

Existing benchmarks~\citep{bellemare2013ale,gym_minigrid,tassa2018dmcontrol} for \gls{rl} do not test the capability of agents in both partial observability and stochasticity. The Atari~\cite{bellemare2013ale} benchmark comprises of 55 games but most of the games are deterministic and a lot of compute is required to train on them. Some tasks in the Atari and Minigrid benchmarks are partially-observable but either lack stochasticity or are hard exploration tasks. Also, these benchmarks do not allow for controlling the factors of variation. We developed a new partially-observable and stochastic environment, called \gls{bbs}, where the task is to push objects to a predefined goal area. Solving tasks in \gls{bbs} require agents to remember states of previously observed objects and avoid noisy distractor objects. Furthermore, \alg~and \algsimple~outperformed leading model-based and model-free baselines. We also present studies with varying partial-observability and stochasticity to demonstrate that the proposed agents have better memory for tracking observed objects and are more robust to increasing levels of noise. Lastly, the proposed methods were also evaluated on existing benchmarks - DeepMind Control~(DMC)~\citep{tassa2018dmcontrol}, DMC with Natural Background~\citep{zhang2021learning,pmlr-v139-nguyen21h}, and Atari~\cite{bellemare2013ale}. On the existing benchmarks, the proposed method performed better on tasks with changing viewpoints and sparse rewards. 

Our key contributions are summarized as follows:
\begin{itemize}
    \item \textbf{Variational Sparse Gating}: We introduce \acrfull{alg}, where the recurrent states are sparsely updated through a stochastic gating mechanism. A comprehensive empirical evaluation shows that \gls{alg} outperforms baselines on tasks requiring long-term memory.
    \item \textbf{Simple Variational Sparse Gating}: We also propose \acrfull{algsimple} which has a purely stochastic state, and achieves competitive results on continuous control tasks when compared with agents that also use a deterministic component.
    \item \textbf{BringBackShapes}: We developed the \glsfirst{bbs} environment to evaluate agents on partially-observable and stochastic settings where these variations can be controlled. Our experiments show that the proposed agents are more robust to such variations.
\end{itemize}
\section{Variational Sparse Gating}
\label{sec:vsg}
\begin{figure}[t]
\minipage{0.46\textwidth}
\centering
\includegraphics[width=\linewidth]{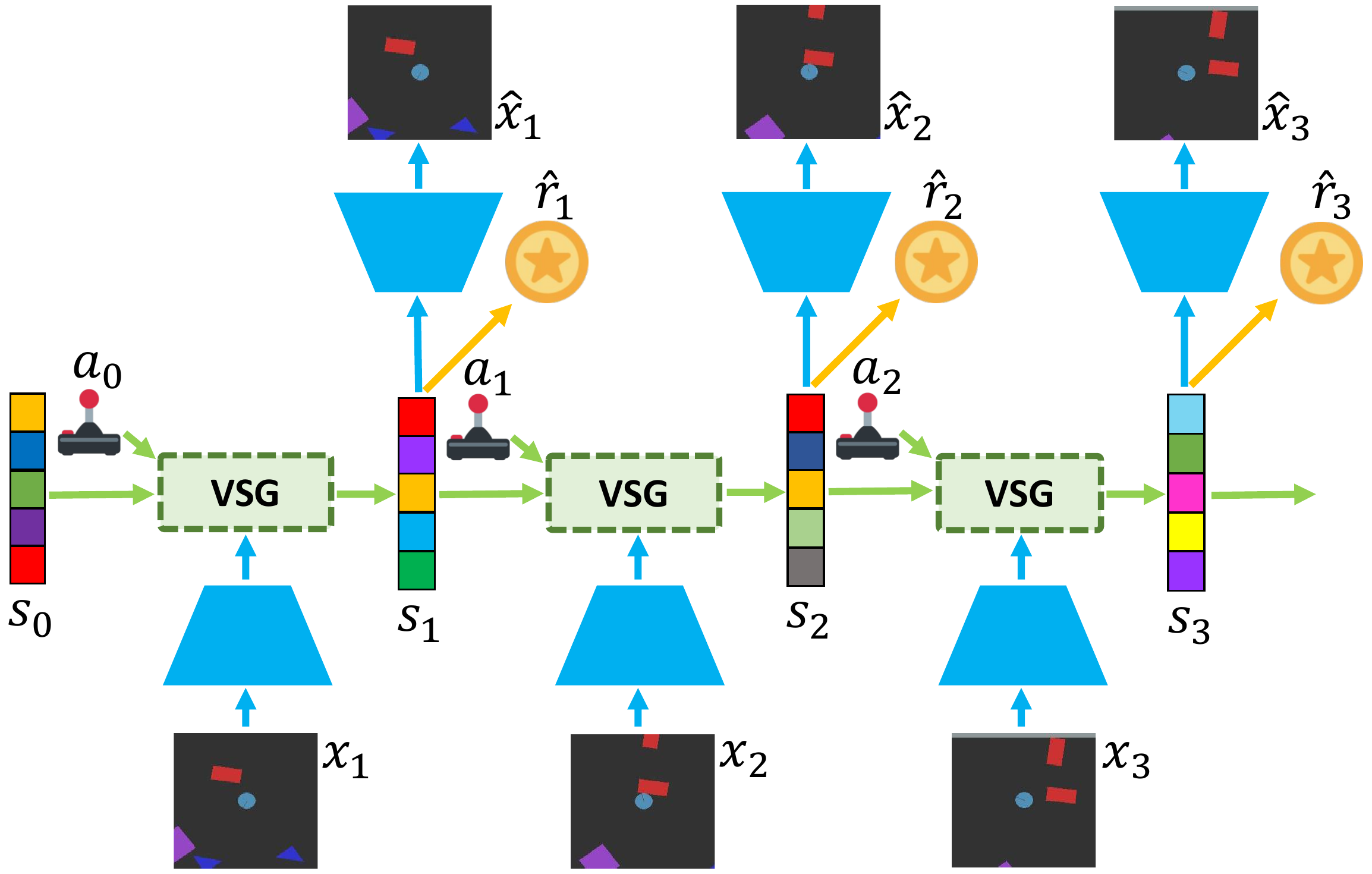}
  \subcaption{} 
  \label{fig:WM}
\endminipage\hfill
\minipage{0.46\textwidth}
\centering
  \includegraphics[width=\linewidth]{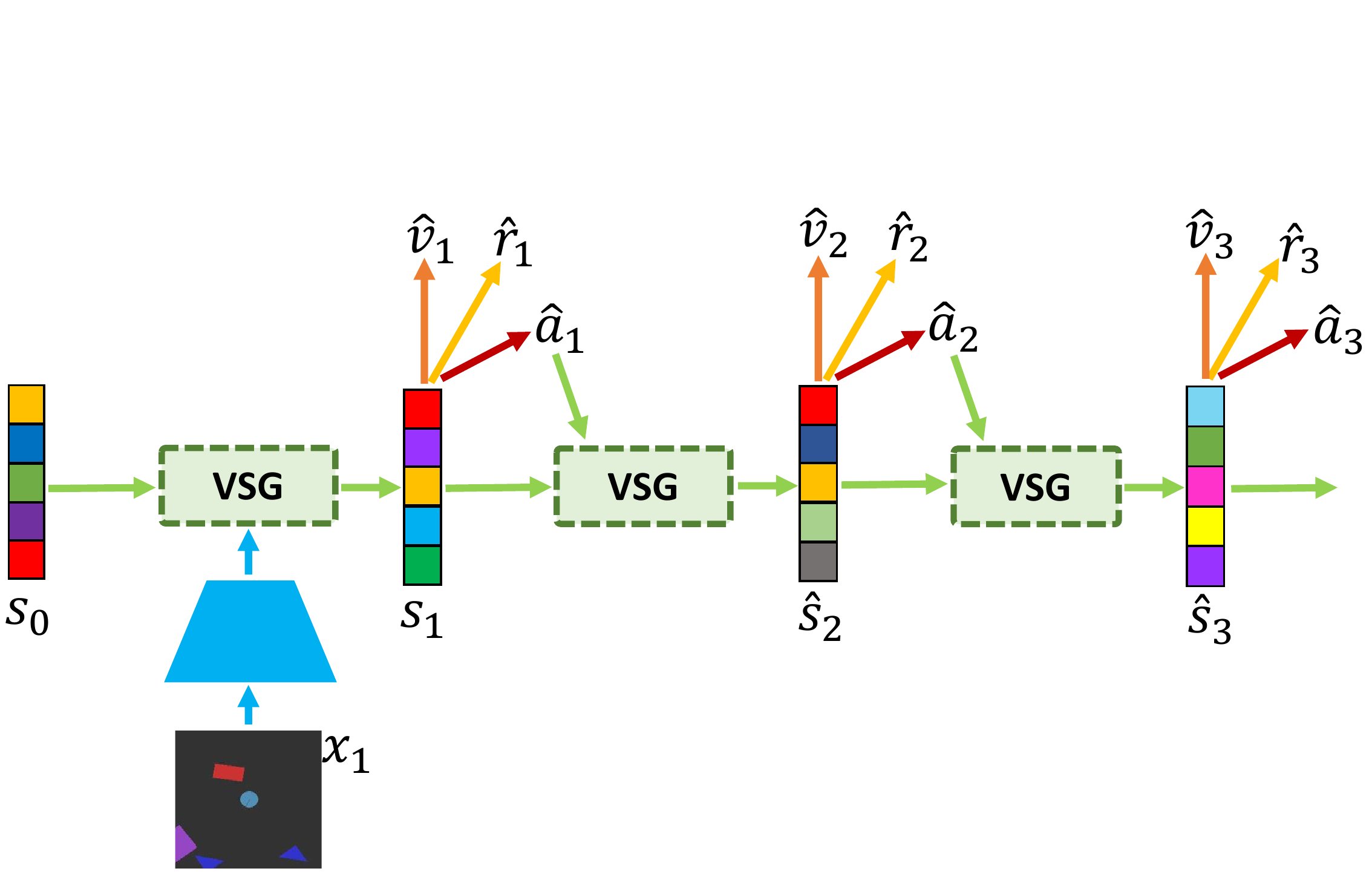}
  \subcaption{}
  \label{fig:BL}
\endminipage
\caption{(a) World Model: The \gls{alg} block takes the previous model state $s_{t-1}$ and action $a_{t-1}$, and outputs the updated model state at next step $s_t$, which is further used to reconstruct image~$\hat{x}_t$ and reward~$\hat{r}_t$. (b) Policy: Comprises of an actor to select optimal action~$\hat{a}_t$ and critic to predict value~$\hat{v}_t$ beyond the planning horizon. The world model is unrolled using the prior model state~$\hat{s}_t$ which does not contain information about image~$x_t$.} 
\label{fig:vsg}
\end{figure}

\textbf{Reinforcement Learning}: The visual control task can be formulated as a Partially Observable Markov Decision Process (POMDP) with discrete time steps $t \in [1;T]$. The agent selects action $a_t \sim p(a_t | o_{\leq t}, a_{<t}$) to interact with the environment and receives the next observation and scalar reward $o_t,r_t \sim p(o_t,r_t|o_{<t},r_{<t}$), respectively, at each time step. The goal is to learn a policy that maximizes the expected discounted sum of rewards $\mathbb{E}_p(\sum_{t=1}^T \gamma^{t} r_t)$, where $\gamma$ is the discount factor. 

\textbf{Agent}: Agent is composed of a world model and a policy (Fig.~\ref{fig:vsg}). World models~(Sec.~\ref{sec:world_model}) encode a sequence of observations and actions into latent representations. The agents behavior~(Appendix~\ref{app:behavior_learning}) is derived to maximize expected returns on the trajectories generated from the learned world model. While training, the world model is learned with collected experience, the policy is improved on trajectories unrolled using the world model and new episodes are collected by deploying the policy in the environment. An initial set of episodes are collected using a random policy. As training progresses, new episodes are collected using the latest policy to further improve the world model. 

\subsection{World Model}
\label{sec:world_model}
\begin{figure}[t]
\minipage{0.30\textwidth}
\centering
\includegraphics[width=\linewidth]{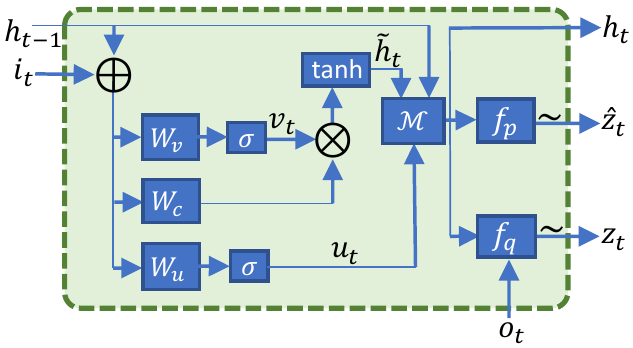}
  \subcaption{}\label{fig:RSSM}
\endminipage\hfill
\minipage{0.30\textwidth}
\centering
  \includegraphics[width=\linewidth]{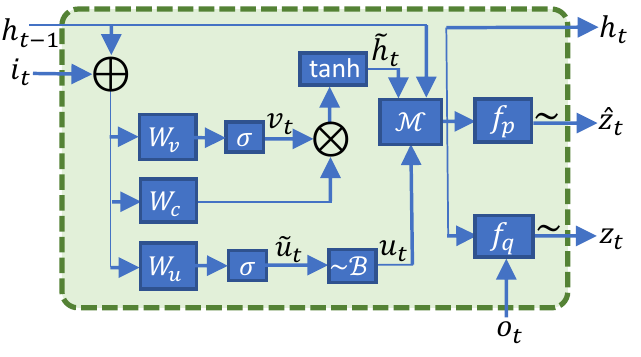}
  \subcaption{}\label{fig:VSG}
\endminipage\hfill
\minipage{0.31\textwidth}%
\centering
  \includegraphics[width=\linewidth]{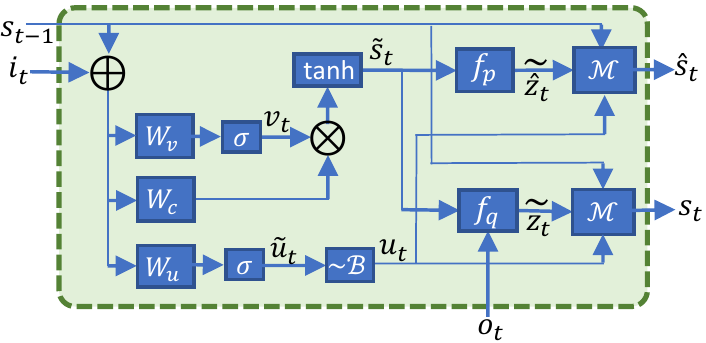}
  \subcaption{}\label{fig:SVSG}
\endminipage
\caption{Architectures of (a) \glsfirst{rssm}, (b) \glsfirst{alg}, and (c) \glsfirst{algsimple}, respectively. $\sigma$ and $\operatorname{tanh}$ denote the sigmoid and tanh non-linear activations, respectively. $W_*$ and $b_*$ are the corresponding weights and biases. $\sim$, $\oplus$ and $\otimes$ denote sampling, vector concatenation, and element-wise multiplication, respectively. $\mathcal{M}$ computes $x_t=u_t\tilde{x}_t + (1-u_t) x_{t-1}$, where $x_t=h_t$ is used for RSSM and VSG, and $x_t=s_t$ is used for SVSG. $\mathcal{B}$ denotes Bernoulli distribution. $f_p$ and $f_q$ denote the prior and posterior distributions with learned parameters, respectively (See Appendix~\ref{app:architectures} for more details).}
\label{fig:architectures}
\end{figure}

World Models~\cite{ha2018worldmodels} learn to mimic the environment using the collected experience and facilitate deriving behaviours in the abstract latent space. Given an abstract state of the world and an action, the model applies the learned transition dynamics to predict the resulting next state and reward. \gls{rssm}~\cite{hafner2019planning} was introduced in PlaNet, where the model state was composed of two paths. The recurrent path consists of an \gls{rnn}~(See Figure~\ref{fig:architectures} [a]), and is motivated with reliable long-term information preservation, while the image representation path samples from a learned distribution to account for multiple possible futures~\cite{babaeizadeh2017sv2p}. In this work, we introduce \glsfirst{alg}, where the recurrent path selectively updates a subset of the latent states at each step using a stochastic gating network. Sparse updates enable the agent to have long-term memory and learn robust representations to solve  complex tasks. 

\textbf{Model Components}: The world model comprises of an image encoder, a \alg~model, and predictors for image, discount and reward. The image encoder generates representations~$o_t$ for the observation~$x_t$ using \glspl{cnn}. The \alg~model comprises of a recurrent model equipped with the stochastic gating mechanism to get the recurrent state~$h_t$, and is used to compute two stochastic image representation states. The posterior representation state~$z_t$ is obtained using the representation model and contains information about the current observation~$x_t$. The prior state~$\hat{z}_t$ is obtained from the transition predictor without observing the current observation~$x_t$. This is useful while planning as sequences are generated in compact latent state, and the output from the transition predictor is utilized. This also results in a lower memory footprint and enables predictions of thousands of trajectories in parallel on a single GPU. The representation states are sampled from a known distribution with learned parameters like Gaussian~\citep{hafner2020dreamer} or Categorical~\citep{hafner2021mastering}. The concatenation of outputs from the recurrent and image representation models gives the compact model state~($s_t=[h_t, z_t]$). The posterior model state is further used to reconstruct the original image~$\hat{x}_t$, predict the reward~$\hat{r}_t$, and discount factor~$\hat{\gamma}_t$. The discount factor helps to predict the probability that an episode will end. The components of the world model are as follows:
\eq{
\begin{alignedat}{4}
& \text{Recurrent model:}        \padspace && h_t            &\ =    &\ f_\phi(h_{t-1},z_{t-1},a_{t-1}) \\
& \text{Representation model:}   \padspace && z_t            &\ \sim &\ \qp(z_t | h_t,x_t) \\
& \text{Transition predictor:}   \padspace && \hat{z}_t      &\ \sim &\ \pp(\hat{z}_t | h_t) \\
& \text{Image predictor:}        \padspace && \hat{x}_t      &\ \sim &\ \pp(\hat{x}_t | h_t,z_t) \\
& \text{Reward predictor:}       \padspace && \hat{r}_t      &\ \sim &\ \pp(\hat{r}_t | h_t,z_t) \\
& \text{Discount predictor:}     \padspace && \hat{\gamma}_t &\ \sim &\ \pp(\hat{\gamma}_t | h_t,z_t).
\end{alignedat}}

\textbf{Neural Networks}: The representation model outputs the posterior image representation state~$z_t$ conditioned on the image encoding~$x_t$ and recurrent state~$h_t$. The transition predictor provides the prior image representation state~$\hat{z}_t$. The image encoding~$o_t$ is obtained by passing the image~$x_t$ through \gls{cnn}~\citep{lecun1989cnn} and \gls{mlp} layers. 
In \alg, we propose to modify the \gls{gru} used in RSSM to sparsely update the recurrent state at each step. The model state~$s_t$, which is a concatenation of recurrent and image representation states is passed through several layers of \gls{mlp} to predict the discount and reward, and transposed \gls{cnn} layers are used to reconstruct the image. The \gls{elu} activation is used for training all the components of the world model~\citep{clevert2015elu}.

\textbf{Sparse Gating}: In light of training \glspl{rnn} to capture long-term dependencies, different ways of applying sparse updates have been investigated~\cite{campos2017skip,neil2016phased,goyal2019recurrent}, enabling a subset of state dimensions to be constant during the update. They were found to alleviate the vanishing gradient problem by effectively reducing the number of sequential operations~\citep{campos2017skip}. Discrete gates may also improve long-term memory by avoiding the gradual change of state values introduced by repeated multiplication with continuous gate values in standard recurrent architectures. Previous works on sparsely updating hidden states~\cite{campos2017skip,neil2016phased} use a separate layer applied over the outputs of \gls{rnn}, and do not modify the \gls{rnn} in itself. However, in this work, we modify the update gate in \gls{gru}~\cite{cho2014gru} to take binary values by sampling from a Bernoulli distribution~(Fig.~\ref{fig:architectures} [b] shows the architecture). 

The input~$i_t$ to the recurrent model contains information about the action and is obtained by concatenating the previous image representation state~$z_{t-1}$ and action~$a_t$ followed by passing them through a MLP layer. Similar to \gls{gru}~\citep{cho2014gru}, there is a reset and update gate.  The reset gate~$v_t$ decides the extent of information flow from the previous recurrent state and inputs, and the update gate~$u_t$ tells which parts of the recurrent state will be updated. The update gate takes only binary values, selecting whether the value will be updated or copied from previous time step. Binary values are obtained by sampling from a Bernoulli distribution where the probability of sampling is obtained using the previous recurrent state~$h_{t-1}$ and input~$i_t$. Straight-through estimators~\cite{bengio2013straight} were used for propagating gradients backwards for training. 
The update equations are:
\eq{
    v_t &= \sigma(W_v^T [h_{t-1}, i_t] + b_v)\\
    \tilde{u}_t &= \sigma(W_u^T [h_{t-1}, i_t] + b_u)\\
    \tilde{h}_t &= \operatorname{tanh}(v_t * (W_c^T [h_{t-1}, i_t] + b_c))\\
    u_t &\sim \operatorname{Bernoulli}(\tilde{u}_t)\\
    h_t &= u_t \odot \tilde{h}_t + (1 - u_t) \odot h_{t-1},
}
where $\odot$ denotes element-wise multiplication, $\sigma$ and $\operatorname{tanh}$ are the sigmoid and hyperbolic tangent activation function, and $W_{*}$ and $b_{*}$ denotes the weights and biases, respectively. To control the sparsity of updates, we have used KL divergence between probability of sampling the update gate~$\tilde{u}_t$ and a fixed prior probability~$\kappa$, where $\kappa$ is a tunable hyperparameter. 

\textbf{Loss function}: The predictors for image and reward produces Gaussian distributions with unit variance, whereas the discount predictor predicts a Bernoulli likelihood. The image representation states are sampled from a Gaussian~\citep{hafner2020dreamer} or a Categorical~\citep{hafner2021mastering} distribution which are trained to maximize the likelihood of targets. In addition, there is a KL Divergence term between the prior and posterior distributions and similar to DreamerV2~\citep{hafner2021mastering}, we have also used KL balancing with a factor of 0.8. We have also added a sparsity loss to regularize the number of updates in hidden state at each step. 
All the components of the world model are optimized jointly using the loss function given by:
\eq{
&\mathcal{L}(\phi)
\doteq
\operatorname{E}_{\qp(z_{1:T} | a_{1:T}, x_{1:T})}\Big[
  \textstyle\sum_{t=1}^T
    \describe{-\ln\pp(x_t | h_t,z_t)}{image log loss}
    \describe{-\ln\pp(r_t | h_t,z_t)}{reward log loss} \\
    &\ \ \describe{-\ln\pp(\gamma_t | h_t,z_t)}{discount log loss} 
    \describe{+\beta\KL[\qp(z_t | h_t,x_t) || \pp(z_t | h_t)]}{KL loss}
\describe{+\alpha\KL[\tilde{u}_t || \kappa]}{sparsity loss}
\Big],
\label{eq:model_loss}\raisetag{11ex}}

where $\beta$ and $\alpha$ are the scale for KL losses of the latent codes and the sparse update gates, respectively.

\section{Simple Variational Sparse Gating}
\label{sec:svsg}
\gls{ssm} were proposed in PLaNet~\cite{hafner2019planning}, where it was discussed that it is not trivial to achieve competitive results without the deterministic recurrent path. Having a deterministic component was motivated to allow the transition model to retain information for multiple time steps as the stochastic component induces variance~\citep{hafner2019planning}. In this work,  we show that having a purely stochastic component achieves comparable performance with DreamerV2 while significantly outperforming SSMs~(refer to Appendix~\ref{app:ssm} for more details). We introduce a simplified version of \alg, called \glsfirst{algsimple}~where the world model has a model state with single path to preserve information over multiple steps and also account for partial observability in future states~(Fig.~\ref{fig:architectures} [c] presents the \algsimple~architecture). 

\textbf{Model Components}: In \algsimple, there is no recurrent model and the posterior state~$s_t$ is obtained using the representation model by conditioning on the previous state~$s_{t-1}$, input image~$x_t$ and the action~$a_t$. Similar to \alg, there is a transition predictor that returns the prior state~$\hat{s}_t$ which does not use the current image observation to imagine trajectories in the latent space. Both the modules sparsely update the model state at each step using the stochastic gating mechanism proposed in \alg. We have used a Gaussian distribution for the stochastic state with a learnable mean vector and a learnable diagonal covariance matrix. Similar to \alg, the posterior state is used to reconstruct the image, and predict the reward and discount factor. The components of world model in \gls{algsimple} are:
\eq{
\begin{alignedat}{4}
& \text{Representation model:}   \padspace && s_t            &\ \sim &\ \qp(s_t | s_{t-1},x_t,a_t) \\
& \text{Transition predictor:}   \padspace && \hat{s}_t      &\ \sim &\ \pp(\hat{s}_t | s_{t-1},a_t) \\
& \text{Image predictor:}        \padspace && \hat{x}_t      &\ \sim &\ \pp(\hat{x}_t | s_t) \\
& \text{Reward predictor:}       \padspace && \hat{r}_t      &\ \sim &\ \pp(\hat{r}_t | s_t) \\
& \text{Discount predictor:}     \padspace && \hat{\gamma}_t &\ \sim &\ \pp(\hat{\gamma}_t | s_t).
\end{alignedat}
}

The representation model~$q_{\phi}$ and transition predictor~$p_{\phi}$ are modified to output the posterior~$s_t$ and prior~$\hat{s}_t$ states, respectively. The reset gate~$v_t$ and the update gate~$\tilde{u}_t$ is calculated using the previous state~$s_{t-1}$ and input~$i_t$ which has the information about the action~$a_t$. The candidate state~$\tilde{s}_t$ at each step is obtained using input $i_t$, reset gate $v_t$ and previous state $s_{t-1}$. Similar to \alg, the update gate $u_t$ is sampled from a Bernoulli distribution to sparsely update the latent states at each step, given by:
\eq{
    v_t &= \sigma(W_v^T [s_{t-1}, i_t] + b_v)\\
    \tilde{u}_t &= \sigma(W_u^T [s_{t-1}, i_t] + b_u)\\
    \tilde{s}_t &= \operatorname{tanh}(v_t \odot (W_c^T [s_{t-1}, i_t] + b_c))\\
    u_t &\sim \operatorname{Bernoulli}(\tilde{u}_t),
}
where $\odot$ denotes the element-wise multiplication, $\sigma$ and $\operatorname{tanh}$ are the sigmoid and hyperbolic tangent activation functions, and $W_{*}$ and $b_{*}$ denote the weights and biases, respectively.

The candidate state~$\tilde{s}_t$ is feeded through \gls{mlp} layers to get the prior and posterior distributions. The image encoding~$x_t$ was used to get posterior distribution, whereas the prior distribution was predicted without it. The prior~$\hat{z}_t$ and posterior~$z_t$ candidate states are sampled from these distributions, where the update gate sparsely modifies the previous latent state and outputs the prior~$\hat{s}_t$ and posterior~$s_t$ model states at each step, respectively. The update equations are given by:
\eq{
    \hat{z}_t &\sim f_{\mathrm{p}}(\tilde{s}_t)\\
    z_t &\sim f_{\mathrm{q}}(\tilde{s}_t, x_t) \\
    \hat{s}_t &= u_t \odot \hat{z}_t + (1 - u_t) \odot s_{t-1} \\
    s_t &= u_t \odot z_t + (1 - u_t) \odot s_{t-1},
}
where $f_{\mathrm{p}}$ and $f_\mathrm{q}$ denotes functions that output a distribution with learnable parameters for prior and posterior, respectively. For SVSG, Categorical latents~\citep{hafner2021mastering} were not performing well on our tasks. We attribute this to the fact that samples from a categorical distribution are binary vectors and it is difficult to accurately reconstruct with such sparse latent representations. Lastly, we observed that sparse gating mechanism introduced in \alg~was important for convergence of \algsimple. 

\textbf{Loss function}: We have used the same loss function as described in Sec.~\ref{sec:vsg} and policy is similar to used in \gls{alg} (described in Appendix~\ref{app:behavior_learning}). For training the \algsimple~model, we replace the KL loss term between prior and posterior distributions in Eq.~\ref{eq:model_loss} with a masked KL loss that penalizes the state dimensions that were updated in the corresponding time step, i.e. those for which the corresponding element in $u_t$ is equal to 1. We found this to be necessary, since the original, unmasked KL loss did not yield good performance, presumably due to its effect on state dimensions that were not updated. 

\section{Experiments}
\label{sec:experiments}
\subsection{\env}
\label{sec:shapeherd}
\begin{figure*}
\centering
\begin{minipage}{.25\textwidth}
  \centering
    \includegraphics[width=0.95\textwidth]{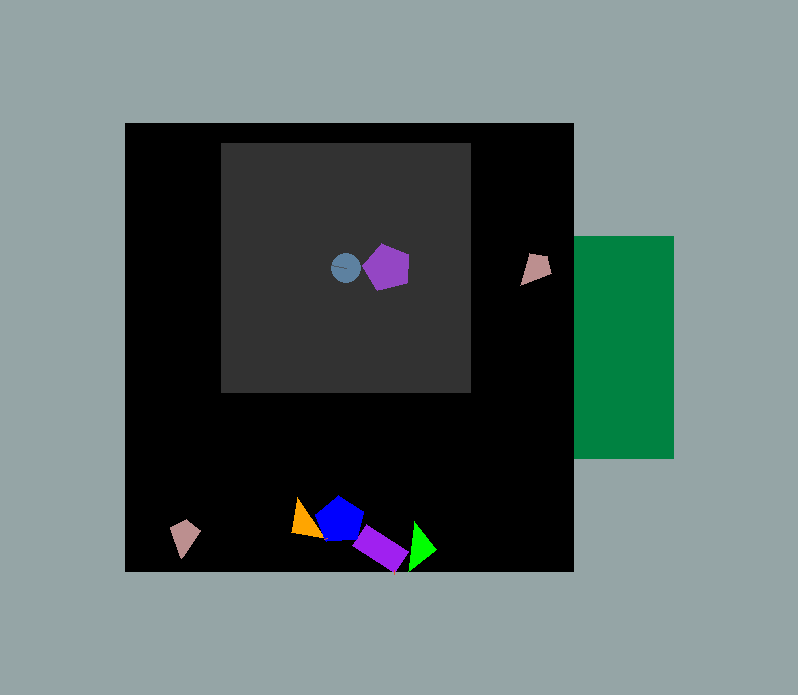}\\
    \includegraphics[width=0.95\textwidth]{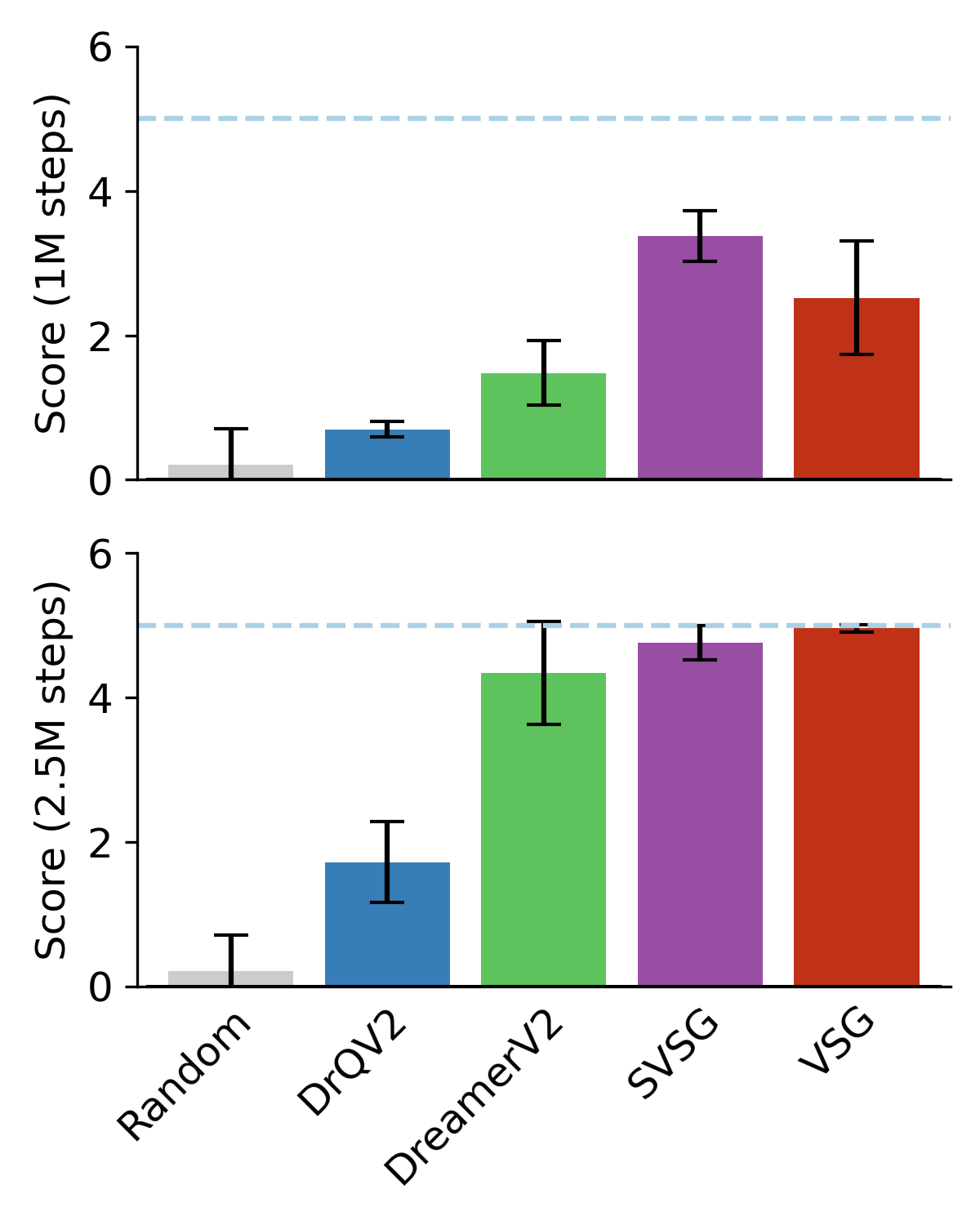}
  \label{fig:shapeherd}
\end{minipage}%
\begin{minipage}{.75\textwidth}
  \centering
  \includegraphics[width=0.9\textwidth]{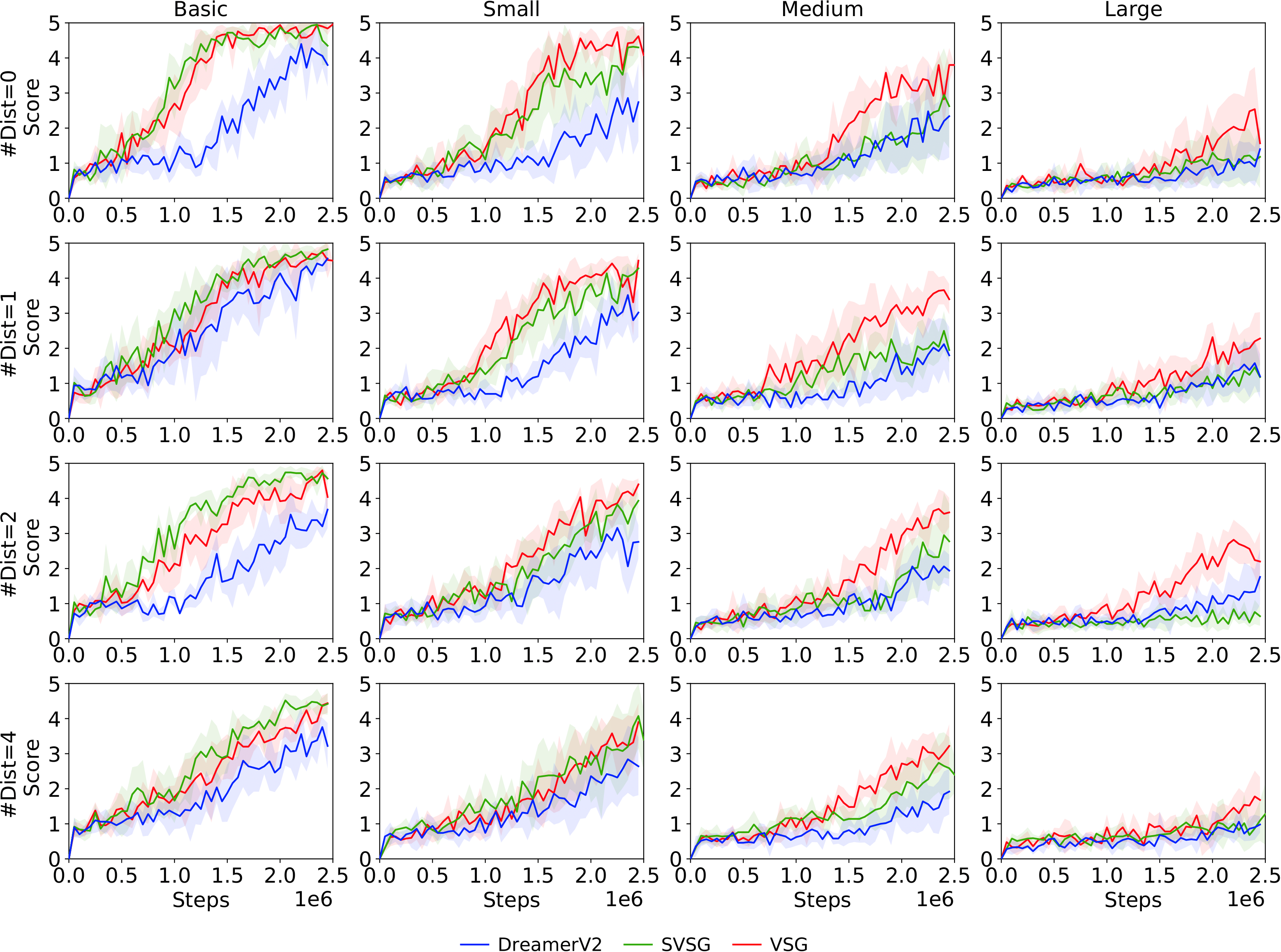}
  \label{fig:shapeherd_results}
\end{minipage}
\caption{a) (Top Left) Full arena of the \glsfirst{bbs}~where the gray region around agent shows the partial view received by it. The circular agent is located in the center of partial view and is of teal blue color. The task is to push objects in the green goal region on the right side of arena. b) (Bottom left) Scores obtained on BBS with Basic size and no distractors at 1M and 2.5M steps. c) (Right) Performance (results over 5 seeds are reported) at different sizes of arena and number of distractor objects~(\#Dist). \gls{alg} and \gls{algsimple} outperforms DreamerV2 significantly in most scenarios. }
\label{fig:bbs_first}
\end{figure*}

\textbf{Environment}: In this work, we developed the \glsfirst{bbs} environment to test the ability of agents to solve tasks in partially-observable and stochastic scenarios~(see Fig.~\ref{fig:bbs_first} [a]). The task is to push the objects within the arena into a pre-specified goal area. Moreover, rewards are sparse and is +1 for successfully pushing an object into the goal. At each time step, the agent only receives an obfuscated view of the arena centered around its current position. This requires agents to efficiently explore the arena to find new objects as well as remember states of previously observed objects. The objects can collide with each other and the walls, which further requires the agent to account for these events while updating its state. Stochasticity was introduced in the environment by using random distractor objects. The distractor objects follow Brownian motion in any direction and can't be pushed into the goal area. Additionally, they add noise to the reward signal as they might push objects into the goal, causing a reward that is not or only partially related to the agent's behavior. Due to partial-observability, such instances might not be visible to the agent, making the task even more challenging. Refer to Appendix~\ref{app:bbs} for further description of the environment.

\begin{figure*}[t]
    \centering
    \includegraphics[width=\textwidth]{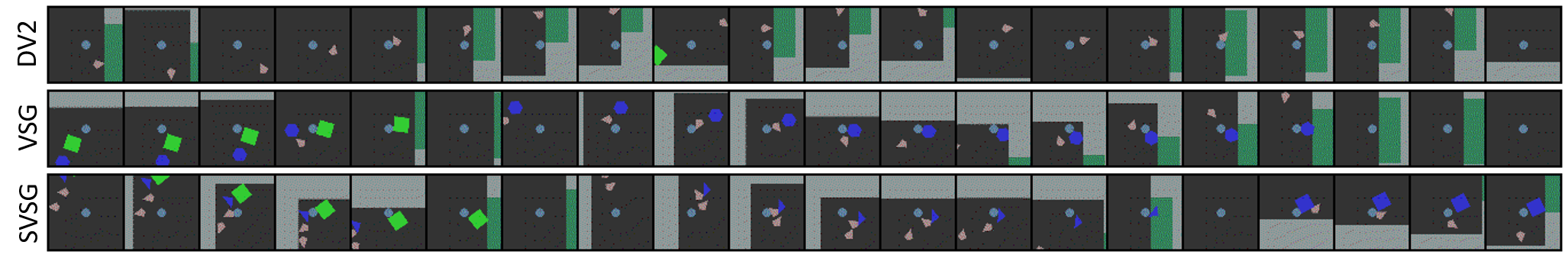}
    \caption{Learned behaviors of DreamerV2~(DV2), VSG and SVSG agents on \env~on Basic size and with 2 distractor objects at different steps. DreamerV2 fails to capture that distractors~(whitish cones) are noisy objects and tries to push them towards the goal, whereas \alg~and \algsimple~learn to avoid the noisy objects and carefully maneuvers the right objects towards the goal.}
    \label{fig:behavior_comparison}
\end{figure*}
\begin{table}[t!]
	\centering
	\scriptsize
		\begin{tabular}{l|cccc|cccc}
			\toprule
             & \textbf{\begin{tabular}[c]{@{}c@{}}First-Visit \\ Time \end{tabular}} & \textbf{\begin{tabular}[c]{@{}c@{}}Episode \\ Length \end{tabular}} & \textbf{\begin{tabular}[c]{@{}c@{}}Objects Not \\ Visited (\%) \end{tabular}} & \textbf{\begin{tabular}[c]{@{}c@{}}Visited Objects \\ Not Scored (\%) \end{tabular}} &
             \textbf{\begin{tabular}[c]{@{}c@{}}First-Visit \\ Time \end{tabular}} & \textbf{\begin{tabular}[c]{@{}c@{}}Episode \\ Length \end{tabular}} & \textbf{\begin{tabular}[c]{@{}c@{}}Objects Not \\ Visited (\%) \end{tabular}} & \textbf{\begin{tabular}[c]{@{}c@{}}Visited Objects \\ Not Scored (\%) \end{tabular}} \\
             \midrule \midrule
			 & \multicolumn{4}{c}{Basic, \#Distractors=0} & \multicolumn{4}{c}{Medium, \#Distractors=0}\\
			 \midrule
              DV2 & 500.25   & 2503.55  & 4.6      & 14.67  & 1800.79  & 2994.91  & 50.2      & 33.68 \\
VSG       & \textbf{276.20}   & \textbf{1881.80}  & \textbf{0.08}      & \textbf{1.38} & \textbf{1360.32}  & \textbf{2953.21}  & 32.6 & \textbf{24.98}\\
SVSG      & 365.08   & 2196.37  & 2.00      & 5.15 & 1375.99  & 2964.85  & \textbf{30.00} & 38.53 \\ 
              \midrule
              & \multicolumn{4}{c}{Basic, \#Distractors=2} & \multicolumn{4}{c}{Medium, \#Distractors=2}\\
              \midrule
              DV2 & 593.53   & 2908.73  & 6.40      & 35.90 & 1669.42  & 2998.51  & 40.80      & 45.48\\
              VSG       & 330.74   & 2482.40  & \textbf{0.4}      & 9.67 & \textbf{1051.29}  & \textbf{2944.34}  & \textbf{16.4} & \textbf{32.12} \\
              SVSG      & \textbf{313.95}   & \textbf{2292.00}  & 0.6      & \textbf{7.61} & 1484.09  & 2988.79  & 31.80 & 51.42 \\
            \bottomrule
    	\end{tabular}
	\caption{ Average values of the first-visit time, episode length, \% of objects not visited, and \% of objects visited but not scored within an episode for trained agents on Basic and Medium environments, and with 0 and 2 distractor objects respectively. Metrics were calculated for 50 episodes for 5 seeds. \alg~and \algsimple~significantly outperformed DreamerV2 on most statistics.}
	\label{tab:discovery_metrics}
\end{table}

\textbf{Experimental Setup}: The~\glsfirst{bbs} environment returns high dimensional images of shape $64\times64\times3$ as observation. Action is a 2-dimensional continuous vector with acceleration and direction as components. Episodes last for 3000 environment steps and an action repeat~\cite{mnih2016a3c} of 4 was used. Baseline agents include DreamerV2~\citep{hafner2021mastering} and DrQ-v2~\cite{yarats2022mastering}. In Appendix~\ref{app:hyperparameter}, we mention the hyperparameters for the proposed methods- \alg~and \algsimple. The model was implemented using Tensorflow Probabability~\cite{dillon2017tensorflow} and trained on a single NVIDIA V100 GPU with 16GB memory. Training time for DreamerV2, VSG and SVSG methods on the BBS environment for 2.5M environment steps are around 12, 11 and 10.5 hours, respectively. Lastly, results are reported across 5 seeds.  \footnote{Code is available at: \url{https://github.com/arnavkj1995/VSG}.}

\textbf{Quantitative Results}: Fig.~\ref{fig:bbs_first}~[b] compares the proposed methods \alg~and \algsimple~with the leading \gls{rl} agents- DreamerV2~\citep{hafner2021mastering} and DrQ-V2~\citep{yarats2022mastering}. The score indicates how many objects on average were scored within a episode. As discussed in Section~\ref{sec:svsg}, we trained \algsimple~with Gaussian latents only. Upon evaluation at 2.5M timesteps, DreamerV2 achieves competitive scores when compared to the proposed methods. Whereas at 1M steps, DreamerV2 does not perform as well as \alg, which has mean score of \textbf{4.9}. This shows that learning with sparsity priors helps improve the sample efficiency. Furthermore, performance of \algsimple~is better than DreamerV2 but similar to \alg, demonstrating that a purely stochastic model can achieve similar performance.

\textbf{Varying Partial-Observability and Stochasticy}: We also study the effect of partial observability and stochasticity. For partial-observability, we increased the size of the arena while reducing the portion visible to the agent. We consider 4 configurations of the arena- Basic, Small, Medium and Large. For stochasticity, we increase the number of distractor objects using values 0, 1, 2, and 4. Fig.~\ref{fig:bbs_first}~[c] presents the plots of models trained at 2.5M steps at different sizes of the arena and number of distractor objects. It can be observed that increasing the size of arena makes it harder to score objects. \alg~was found to outperform DreamerV2 across all arena sizes. However, \algsimple~did not perform well on larger arena sizes. Adding noisy distractor objects led to drop in final performance of all models. But \alg~and \algsimple~still outperformed DreamerV2, indicating that the sparsity prior helps in ignoring the noisy objects in the arena while solving the task.  

\textbf{Ablation Studies}: In this work, we also report statistics to describe the behavior of learned agents. First-visit time is the number of episode steps taken to visit an object~(when object is completely visible in the agent's view) and is calculated by averaging the first-visit time of each object in the arena. Lower first-visit time indicates that an agent is able to quickly discover all the objects in the arena. Another metric is Episode Length which denotes the number of steps taken by the agent to complete the task. The maximum of these scores was set to 3000. We also report the percentage of objects that were visited within an episode which represents the ability of agents to explore all parts of the arena to find novel objects. Lastly, we report the percentage of visited objects that were not scored which indicates us that the agent might not be remembering positions of objects seen previously. Thus, agents might have to explore the arena again to find them leading to an increase in time taken to finish the task.  Table~\ref{tab:discovery_metrics} presents the results on different settings of the environment and it can be observed that \gls{alg} and \gls{algsimple} outperform DreamerV2 significantly. 

\textbf{Qualitative Results}: We also observed the maneuvers taken by the agents to push the objects to the goal. The DreamerV2 agent was able to recognize objects and go behind them to push, but did not follow a smooth trajectory and was spending more time around an object to push it in the goal. However, our methods showed smoother trajectories and were more efficient at pushing objects successfully in the goal area. Additionally, our methods learned to avoid noisy distractor objects whereas the DreamerV2 agent was colliding with them and was trying to push them to the goal~(see Fig.~\ref{fig:behavior_comparison} and supplementary material for more videos). 

\textbf{Effect of Sparse Gating}: We conducted an experiment where the learned world model was given the first 15 frames and 5 different rollouts were generated in the latent space for the next 35 frames. The sequence of actions is kept fixed across rollouts. The aim was to observe if the sparse gating mechanism is helping the model to retain information for longer time steps and the imagined trajectories are consistent. It was observed in Appendix~\ref{app:latent} that learned world model in VSG and SVSG remembers the color and location of objects, and is also cognizant about the goal location and walls. Furthermore, unrolled trajectories from the world model of DreamerV2 showed distortion in the shapes, and in some instances modifies the color of the objects. 

\subsection{DeepMind Control Suite}
\label{sec:dmc}
\begin{figure*}[t]
    \centering
    \includegraphics[width=0.99\textwidth]{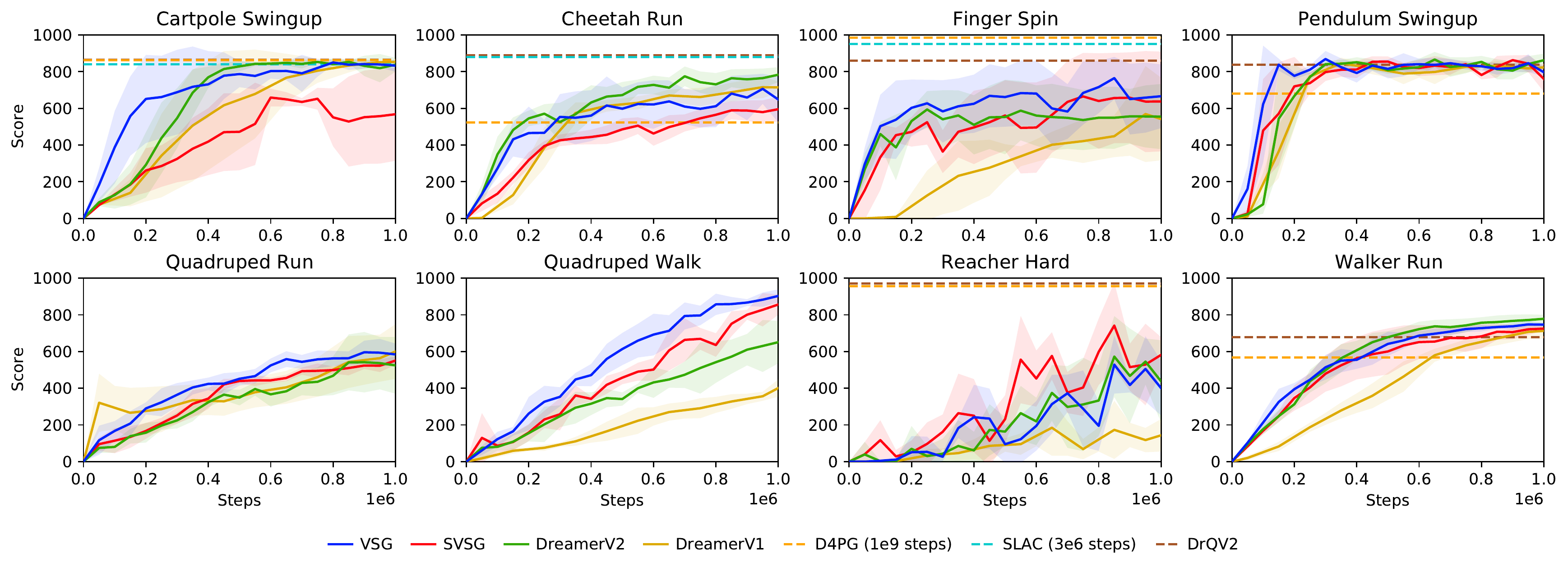}
    \caption{Comparison of \alg~and \algsimple~with DreamerV1~\citep{hafner2020dreamer} and DreamerV2~\citep{hafner2021mastering} on the DeepMind Control Suite. \alg~converges faster on many tasks as demonstrated by evaluation curves. Even with a single stochastic path, \algsimple~achieves performance competitive to the models that use a combination of multiple paths.}
    \label{fig:dmc_results_plots_1M_small}
\end{figure*}

\textbf{Experimental Setup}: The proposed method is evaluated on a few tasks from DeepMind Control Suite~\cite{tassa2018dmcontrol}. Observations for the agents are high dimensional images of shape $64\times64\times3$, actions range between 1 to 12 dimensions, and episodes last for 1000 steps. An action repeat~\cite{mnih2016a3c} of 2 was used. The model was implemented using Tensorflow Probabability~\cite{dillon2017tensorflow} and trained on a single NVIDIA V100 GPU with 16GB memory in less than 6 hours. The agents were trained for 1M environment steps. Baselines include DreamerV1~\citep{hafner2020dreamer}, DreamerV2~\citep{hafner2021mastering}, DrQ-v2~\cite{yarats2022mastering}, D4PG~\citep{barth2018d4pg}, and A3C~\citep{mnih2016a3c}. Except A3C, all baselines learn policies from high dimensional pixel inputs. DreamerV2 was trained using the implementation provided by the authors. For other baselines, the metrics provided by the respective authors were used for comparison. Lastly, returns averaged across 5 seeds were reported.

\textbf{Results}: Figure~\ref{fig:dmc_results_plots_1M_small} presents comparison of \alg~and \algsimple~with the baseline agents~(Refer to Appendix~\ref{app:dmc} for comparison on more tasks). It can be observed that on most tasks, \alg~performs comparable to or better than DreamerV2~\citep{hafner2021mastering}.
Notably, \alg~significantly outperforms DreamerV2 on Quadruped and Finger-Spin tasks. Furthermore, \algsimple~with a purely stochastic component has similar performance to DreamerV2, outperforming on Finger Spin and Quadruped tasks and performing worse on Cartpole Swingup and Cheetah Run tasks. In addition, \algsimple~significantly outperforms DreamerV1 on many tasks which also used Gaussian latents and \gls{rssm} with multiple paths. Lastly, we also present the importance of sparsity loss in \gls{alg}~(See Appendix~\ref{app:ablations_vsg}), and of KL Masking and Sparsity loss in \gls{algsimple}~(see Appendix~\ref{app:ablations_svsg}). 

\textbf{Ablation Studies}: In \gls{bbs}, we added noise in the environment by having distractor objects. We also experimented with other forms of noise where natural videos are used in the background for DeepMind Control tasks~\cite{zhang2021learning,pmlr-v139-nguyen21h}. Since reconstruction-free model based \gls{rl}~\cite{deng2021dreamerpro,nguyen2021temporal} perform better than reconstruction based agents~\cite{hafner2021mastering} in such scenarios, we updated the \gls{rssm} block in DreamerPro~\cite{deng2021dreamerpro} with \gls{alg} and call it VSGPro, and trained it on DMC with natural background~(Refer to Appendix~\ref{app:nat}). We also experimented with \gls{alg} in discrete control tasks from the Atari benchmark in Appendix~\ref{app:atari}, where \gls{alg} performed better on task with changing viewpoints.

\section{Related Work}
\label{Sec:RelatedWork}

\textbf{Latent Dynamics Models}: Latent dynamics models~\cite{bourlard2012connectionist, kalman1960filter, bengio1999markovian} operate directly on sequences predicted in the latent space rather than autoregressively feeding back the generated frames back to the model. Recent advancements in deep learning have allowed learning expressive latent dynamics models using stochastic backpropagation~\cite{kingma2013vae,chung2015vrnn, krishnan2015deepkalman, karl2016dvbf}. \glsfirst{rssm}~\cite{hafner2019planning} comprises of stochastic and deterministic components. VideoFlow~\cite{kumar2019videoflow} predicted future values of the latent state by normalizing flows for robotic object interactions. Hierarchical latent models for video prediction were proposed in CWVAEs~\citep{saxena2021clockwork} with levels ticking at different intervals in time. \cite{franceschi2020stochastic} and \cite{dona2021pdedriven} disentangled dynamic and static factors where 2-5 initial observations was used to estimate the static component. 

\textbf{RL for Visual Control}: Deep Reinforcement Learning (DRL) methods fall into one of two categories: 1) \textit{Model-Based} --- where an explicit model of the environment and its dynamics are learned~\citep{ha2018worldmodels,hafner2019planning,hafner2020dreamer,hafner2021mastering,zhang2018solar,kaiser2019simple}, and 2) \textit{Model-Free} --- where a policy is learned directly from the raw observations~\citep{srinivas2020curl,kostrikov2020image,lillicrap2015ddpg,yarats2022mastering, schwarzer2021dataefficient, pmlr-v162-mondal22a}. Deep Deterministic Policy Gradient (DDPG)~\cite{lillicrap2015ddpg} combined actor-critic with insights from DQNs~\cite{mnih2015dqn} to learn agents for continuous action spaces. TD3~\cite{fujimoto2018td3} builds upon the DDPG algorithm and addresses the problem of overestimation bias in the value function. CURL~\cite{srinivas2020curl} uses contrastive losses to learn discriminative representations. DrQ~\cite{kostrikov2020drq} and  DrQ-v2~\citep{yarats2022mastering} employed data augmentation techniques and do not use auxiliary losses or pre-training. \Gls{rssm} was introduced in PlaNet~\cite{hafner2019planning} and was employed for online planning in the latent space. DreamerV1~\cite{hafner2020dreamer} and DreamerV2~\cite{hafner2021mastering} achieved state-of-the-art results on DMC~\cite{tassa2018dmcontrol} and Atari~\cite{bellemare2013ale}, respectively. SimPLe~\cite{kaiser2019simple} trains a PPO~\cite{schulman2017ppo} agent on the learned video generation model in pixel space. SOLAR~\cite{zhang2018solar} solved robotics tasks via guided policy search. 

\textbf{Sparsity in \gls{rnn}}:
Neural networks have widely adopted sparsity to reduce the memory footprint of weights and activations~\citep{lecun1990optimal,chen2015compressing,han2015deep}. Several works have explored sparsity in \glspl{rnn}. \citet{campos2017skip} introduced a mechanism in \glspl{rnn} that learns to skip state updates, effectively reducing the number of sequential operations on the latent state, thereby alleviating the problem of vanishing gradients in training on long sequences. \citet{goyal2019recurrent} presented \gls{rim}, an architecture that consists of separate recurrent modules which are sparsely updated using a learned attention mechanism. In contrast to \gls{rim}, the number of updated state variables in \alg~algorithm is not fixed.
\section{Discussion}
\label{sec:discussion}
In this work, we introduce \alg~and \algsimple, two latent dynamics models leveraging sparse state updates. The sparse update prior was found to facilitate more efficient behaviors in tasks requiring long-horizon planning. Furthermore, \gls{algsimple} is a purely stochastic model with a single component in the model state. We show that \gls{alg} and \gls{algsimple} can outperform leading agents on the proposed \env~task, a challenging partially-observable and stochastic environment. \gls{bbs} allows for controlling different factors of variation like stochasticity and partial-observability. Experiments conducted on various variations in \gls{bbs} demonstrate that the proposed agents are more robust to noise in the environment and can better retain information of seen objects. Some limitations and potential research directions for future research are as follows:
\begin{itemize}
    \item In the current implementation of \alg, the latent space does not exhibit disentanglement which could be an interesting direction for future research. Gating mechanisms in \alg~can also be combined with other recurrent architectures like \gls{rim}~\citep{goyal2019recurrent}. 
    \item In this work, \gls{bbs} was explored with only 2 factors of variation: partial-observability and stochasticity. More controllable factors like the nature of entities~(shape, size, color of objects), underlying physics~(mass, friction, elasticity), or procedural background generation can be introduced to further study generalization capabilities of \gls{rl} agents. 
    \item \gls{algsimple} being a purely stochastic model can further be used to estimate state uncertainty by marginalizing over multiple samples paths to efficiently explore in an unknown environment. 
    \item  Evaluation on first-person view 3D games like tasks in DMLab~\cite{beattie2016dmlab} would be interesting. Furthermore, a 3D version of the \gls{bbs}~environment with the viewpoints changing with rotation of agent and the underlying physics will make the task more challenging.
    \item We have used small latent dimensions and it would be interesting to train such models with larger architectures and on more complex tasks. Scaling the current architecture would also require optimizing the implementation to make them computationally feasible.
    \item Categorical latents outperformed Gaussian latents as the stochastic states of RSSM~\cite{hafner2021mastering}, especially for discrete control tasks. However, \gls{algsimple} was not found to work well with Categorical latents and we believe that sampled sparse states are hard to optimize.
    \item Model-based \gls{rl} for visual control is still in early stages. However, a major challenge with deploying such models in the real world is safety especially during exploration. This would require an accurate world model that allows learning policies with stringent safety constraints that avoid mistakes when deployed in the real world. Such algorithms will rely on models that are robust when transferred from simulation to the real world.

\end{itemize}

\section*{Acknowledgements}
The authors would like to thank David Meger, Lucas Lehnert and Ankesh Anand for their valuable feedback and discussions. The text also benefited from discussions with Abhinav Agarwalla, Rupali Bhati and Vineet Jain. The authors are also grateful to CIFAR for funding and the Digital Research Alliance of Canada for computing resources.

\bibliographystyle{plainnat}
\bibliography{main}
\section*{Checklist}


\begin{enumerate}

\item For all authors...
\begin{enumerate}
  \item Do the main claims made in the abstract and introduction accurately reflect the paper's contributions and scope?
    \answerYes{}
  \item Did you describe the limitations of your work?
    \answerYes{See Section~\ref{sec:discussion}.}
  \item Did you discuss any potential negative societal impacts of your work?
    \answerYes{Described in the section with limitations.}
  \item Have you read the ethics review guidelines and ensured that your paper conforms to them?
    \answerYes{}
\end{enumerate}

\item If you are including theoretical results...
\begin{enumerate}
  \item Did you state the full set of assumptions of all theoretical results?
    \answerNA{}
        \item Did you include complete proofs of all theoretical results?
    \answerNA{}
\end{enumerate}

\item If you ran experiments...
\begin{enumerate}
  \item Did you include the code, data, and instructions needed to reproduce the main experimental results (either in the supplemental material or as a URL)?
    \answerYes{The code for the model and dataset is in the supplementary material. There is a README file with the instructions to run them.}
  \item Did you specify all the training details (e.g., data splits, hyperparameters, how they were chosen)?
    \answerYes{We have specified the hyperparameters in Appendix~\ref{app:hyperparameter}.}
        \item Did you report error bars (e.g., with respect to the random seed after running experiments multiple times)?
    \answerYes{All the results and plots presented in Section~\ref{sec:experiments} and Appendix were obtained after training on multiple seeds ranging from 3-5. }
        \item Did you include the total amount of compute and the type of resources used (e.g., type of GPUs, internal cluster, or cloud provider)?
    \answerYes{We have mentioned about the GPUs and time taken to run on a single seed in the implementation details for each environment.}
\end{enumerate}

\item If you are using existing assets (e.g., code, data, models) or curating/releasing new assets...
\begin{enumerate}
  \item If your work uses existing assets, did you cite the creators?
    \answerYes{}{}
  \item Did you mention the license of the assets?
    \answerNo{We used the implementation of DreamerV2, DreamerPro and DrQ-v2 provided by the authors with MIT license. For comparison with DBC, we thank the authors for sharing the evaluation logs.}
  \item Did you include any new assets either in the supplemental material or as a URL?
    \answerYes{The code for the new environment (BBS) is included in the supplementary material.}
  \item Did you discuss whether and how consent was obtained from people whose data you're using/curating?
    \answerYes{}
  \item Did you discuss whether the data you are using/curating contains personally identifiable information or offensive content?
    \answerNA{}
\end{enumerate}

\item If you used crowdsourcing or conducted research with human subjects...
\begin{enumerate}
  \item Did you include the full text of instructions given to participants and screenshots, if applicable?
    \answerNA{}
  \item Did you describe any potential participant risks, with links to Institutional Review Board (IRB) approvals, if applicable?
    \answerNA{}
  \item Did you include the estimated hourly wage paid to participants and the total amount spent on participant compensation?
    \answerNA{}
\end{enumerate}

\end{enumerate}


\appendix
\newpage
\section*{Appendix}
\section{BringBackShapes}
\label{app:bbs}
The environment has a circular blue agent which can move in any direction. The shapes and colors of the objects are uniformly sampled from a predefined set with 5 shapes and 5 colors, respectively. As there are 5 objects in the arena for each episode, there are $25^{5}$ $\sim$ 9.8M possible combinations. The initial positions of the agent and objects are randomly chosen within the arena. The elasticity of the objects and the agent is 1.0, while the walls have an elasticity of 0.7. There is a damping factor of 0.3 applied to the velocities of all objects and the agent. In Figure~\ref{fig:bbs_first} [a], we show a full view of the whole arena at the beginning of an episode, and the gray region around the agent is its view. It can be observed that the agent might see none or all of the objects in its view and needs to explore in the environment to look for the objects in order to push them towards the goal (green region in Figure~\ref{fig:bbs_first} [a]). The agent receives image observations of size 64 × 64 × 3 from the environment. The action space is continuous and comprises of the angle and magnitude of force applied by the agent. The rewards are sparse and agent receives a reward of +1 for successfully pushing an object into the goal area. An episode terminates once all objects are pushed into the goal or if 3000 steps are completed. In this work, agents were trained for 2.5M environment steps and an action repeat of 4 was used. Evaluation was performed across 5 seeds with 10 episodes for each seed, and means and standard deviations across the seeds are reported.

Our motivation behind creating BBS was to have a simple benchmark where the factors of variation can be controlled. For instance, in the current version, we add stochasticity and partial-observability. We believe future work can test for generalization to differences in the controllable factors between training and testing. These factors of variation specify the context of the MDP formalism of the task that the agent is trying to solve. If we have control over varying the context, we can define the training and the test distributions and this can enable us to formalize the class of generalisation problems we are focusing on. Furthermore, the environment can also be extended for open-ended learning where the agent has to learn an ever-increasing set of behaviours and abilities. 
Lastly, this can be further extended to multi-agent setting where the behaviour we expect to see is the emergence of some kind of cooperation between agents. 
\section{Behavior Learning}
\label{app:behavior_learning}
The policy is trained by generating trajectories in the latent space obtained from the learned world model. The policy comprises of a stochastic actor and a deterministic critic to learn behaviours in the latent space. The actor learns to choose the most optimal actions conditioned on the model state~($\hat{a}_t \sim \p<p_\psi>(\hat{a}_t | \hat{s}_t)$). The critic estimates the discounted sum of future rewards that are beyond the planning horizon~($v_\xi(\hat{s}_t) \approx \operatorname{E}_{p_\phi,p_\psi}\Big[
  \textstyle\sum_{\tau \geq t} \hat{\gamma}^{\tau-t} \hat{r}_\tau$\Big]).  To obtain the latent trajectories, the initial model state is extracted from the collected data. The actor network provides the action~$\hat{a}_t$, which is used to obtain the prior states~$\hat{z}_t$ at each step. Since the agent does not act using these actions in the environment, the prior distributions are used to sample the state and reward predictor provides the reward~$\hat{r}_t$. Furthermore, the value network provides the discounted sum of future rewards from that state. The actor and critic optimize different objectives:

\textit{Critic Loss}: Temporal Difference learning is used to update the parameters of the critic. The target is estimated by combining the predicted rewards from latent model states and value estimates from critic. The weighted average of n-step returns ($V_{\lambda}$) proposed in DreamerV1~\cite{hafner2020dreamer} is used. The critic parameters ($\xi$) are optimized using the \gls{mse} between the predicted value and the $\lambda$-target over all the states in a trajectory, given by:
\eq{
\mathcal{L}(\xi) \doteq
\operatorname{E}_{p_\phi,p_\psi}\Big[
  \textstyle\frac{1}{H-1} \sum_{t=1}^{H-1}
    \frac{1}{2} \big(
      v_\xi(\hat{s}_t) - \operatorname{sg}(V^\lambda_t)
    \big)^2
\Big],
}
where $sg$ denotes stopping gradients at the target while updating the critic, and $H$ denotes the length of the planning horizon in latent space which was kept to 15 in our experiments. Furthermore, the targets are computed using a copy of the critic which is updated after every 100 gradient steps. 

\textit{Actor Loss}: The actor is trained to maximize the $\lambda$-return  computed for training the critic. The reparameterization trick \cite{hafner2020dreamer, kingma2018glow, rezende2014vae} was used to backpropagate gradients from the value estimate. The entropy of the actor distribution is also regularized to encourage exploration. For training,  $\smash{\eta_d=1.0}$ and the entropy regularizer $\smash{\eta_e=10^{-4}}$ was used. The loss for training the actor parameters ($\phi$) is given by:

\eq{
\mathcal{L}(\psi) \doteq
\operatorname{E}_{p_\phi,p_\psi}\Big[
  \textstyle\frac{1}{H-1} \sum_{t=1}^{H-1} \big(
    \describe{-\eta_d V^\lambda_t}{\ensuremath{\substack{\text{\scriptsize dynamics} \\ \text{\scriptsize backprop}}\vspace*{-2ex}}}
    \describe{-\eta_e \operatorname{H}[a_t|\hat{s}_t]}{entropy regularizer}
\big)\Big].
}
\section{Hyper Parameters}
\label{app:hyperparameter}

\begin{table*}[hbt!]
	\centering
		\begin{tabular}{l|cc}
			\toprule
			\textbf{Name} & \textbf{VSG} & \textbf{SVSG} \\
			\midrule
			\multicolumn{3}{c}{World Model}\\ 
			\midrule
             Batch Size & 16 & 16 \\
             Sequence Length & 50 & 50 \\
             Recurrent state dimensions & 1024 & 1024 \\
             Image Representation num classes & 32 & - \\
             Image Representation class dimension & 32 & -  \\
             KL Loss Scale & 1.0 & 1.0 \\
             KL Balancing & 0.8 & 0.8 \\
             Sparsity Loss Scale & 0.1 & 0.1 \\
             Prior gate probability $\kappa$& 0.3 / 0.4& 0.3 / 0.4 \\
             World Model learning rate & $3\times 10^{-4}$ & $8\times 10^{-4}$ \\
             Reward transformation & Identity & Identity\\
            \midrule
            \multicolumn{3}{c}{Behavior}\\
            \midrule
             Imagination Horizon & 15 & 15 \\
             Discount & 0.99 & 0.99 \\
             $\lambda$-target parameter & 0.95 & 0.95 \\
             Actor Gradient Mixing & 0.1 & 0.1 \\
             Actor Entropy Loss Scale & $1\times 10^{-4}$ / $2\times 10^{-3}$ & $1\times 10^{-4}$ / $2\times 10^{-3}$ \\
             Actor Learning Rate & $8\times 10^{-5}$ & $8\times 10^{-5}$ \\
             Critic Learning Rate & $8\times 10^{-5}$ & $8\times 10^{-5}$ \\
             Slow critic update inverval & 100 & 100\\
            \midrule
            \multicolumn{3}{c}{Common}\\
            \midrule
            Environment steps per update & 5 & 5 \\
            MLP number of layers & 4 & 4 \\
            MLP number of units & 400 & 400\\
            Gradient clipping & 100 & 100\\
            Adam epsilon & $1\times 10^{-5}$ & $1\times 10^{-5}$ \\
            Weight decay & $1\times 10^{-6}$ & $1\times 10^{-6}$\\
            \midrule
            Total Parameters & 32.3M & 30.8M \\
            \bottomrule
    	\end{tabular}
	\caption{Hyper parameters of \alg~and \algsimple. When parameters are separated by /, the left hand side value is for BBS and the right hand side value is for other environments. When tuning the agent for a new task, we recommend searching over the KL loss scale~$\beta \in \{0.1, 0.3, 1, 3\}$, prior gate probability~$\kappa \in \{0.3, 0.4, 0.5\}$ and the discount factor $\gamma \in \{0.99, 0.999\}$. }
	\label{tab:appendix_hyperparameter}
\end{table*}
\newpage
\section{Scores on Deepmind Control Suite}
\label{app:dmc}
\begin{figure}[hbt!]
    \centering
    \includegraphics[width=0.99\columnwidth]{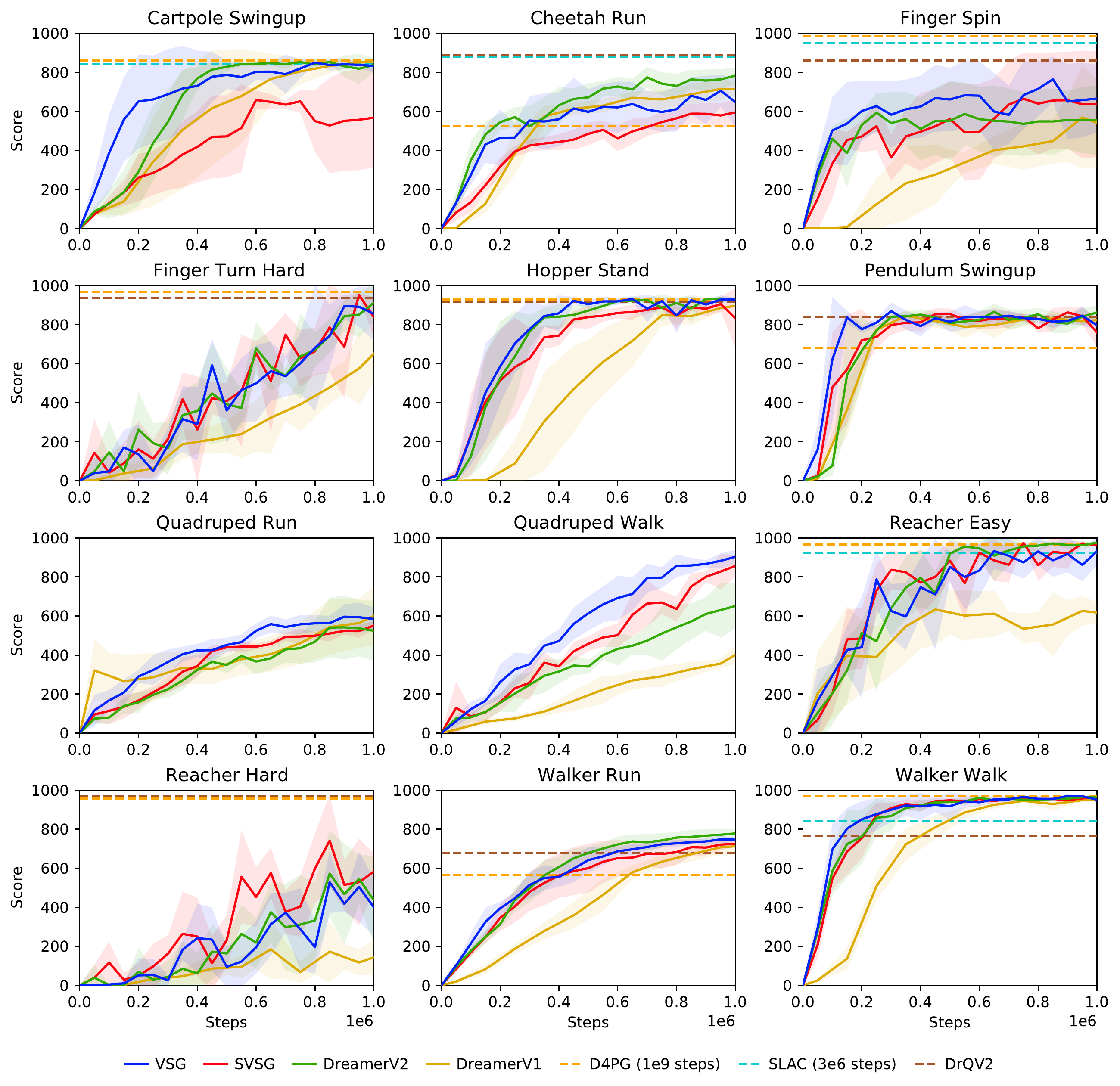}
    \caption{Comparison of \alg~and \algsimple~with leading algorithms like DreamerV1~\citep{hafner2020dreamer}, DreamerV2~\citep{hafner2021mastering} and DrQ-V2~\citep{yarats2022mastering} on tasks from the DeepMind Control Suite.}
    \label{fig:dmc_scores}
\end{figure}

In Fig.~\ref{fig:dmc_scores}, we present the scores on 12 tasks from DMControl Suite. \alg~was found to perform better on 4 tasks and was competitive on 7 tasks when compared with DreamerV2. Furthermore, \algsimple~when compared with DreamerV1 which also used Gaussian latents, was found to perform better on 8 tasks and had similar performance on 2 tasks. This demonstrates that using Gaussian latents with a single path and sparse gating mechanism can achieve competitive results when compared to leading methods and is better than previous methods using Gaussian latents. 

\newpage
\section{Ablations Studies}
In this section, we present ablation experiments on the DeepMind Control Suite \cite{tassa2018dmcontrol}. 

\subsection{Sparsity Loss in \alg}
\label{app:ablations_vsg}
\begin{figure}[hbt!]
    \centering
    \includegraphics[width=0.99\columnwidth]{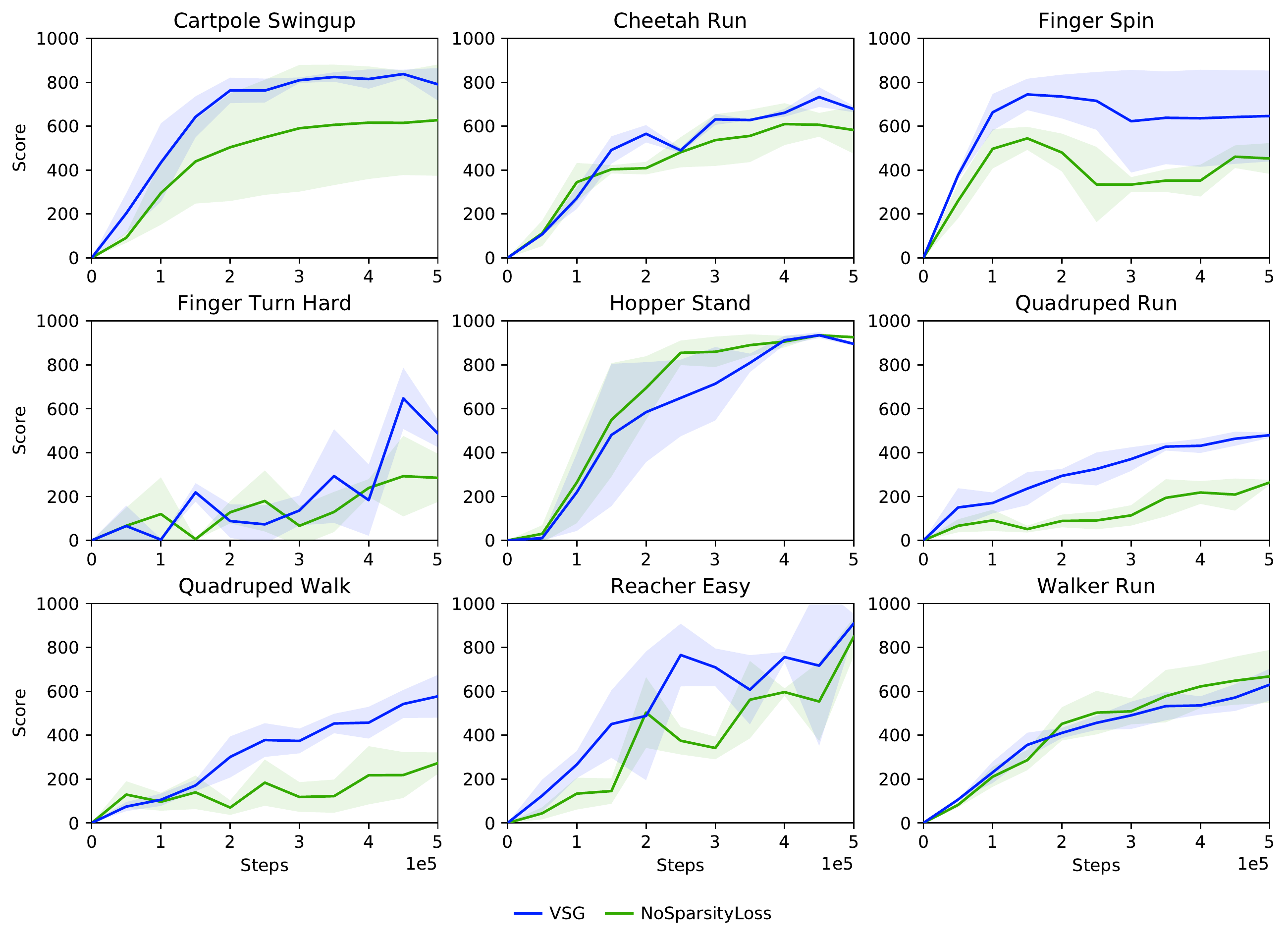}
    \caption{Ablation study showing the performance of VSG on 9 tasks tasks from DMC trained with (\textit{VSG}) and without the sparsity penalty (\textit{NoSparsityLoss}). VSG without the sparsity loss on update gate probabilities was found to significantly underperform on 5 out of 9 tasks from DM Control Suite. }
    \label{fig:ablations_vsg}
\end{figure}

\newpage
\subsection{Sparsity Loss and KL Mask in \algsimple}
\label{app:ablations_svsg}
\begin{figure}[hbt!]
    \centering
    \includegraphics[width=0.99\columnwidth]{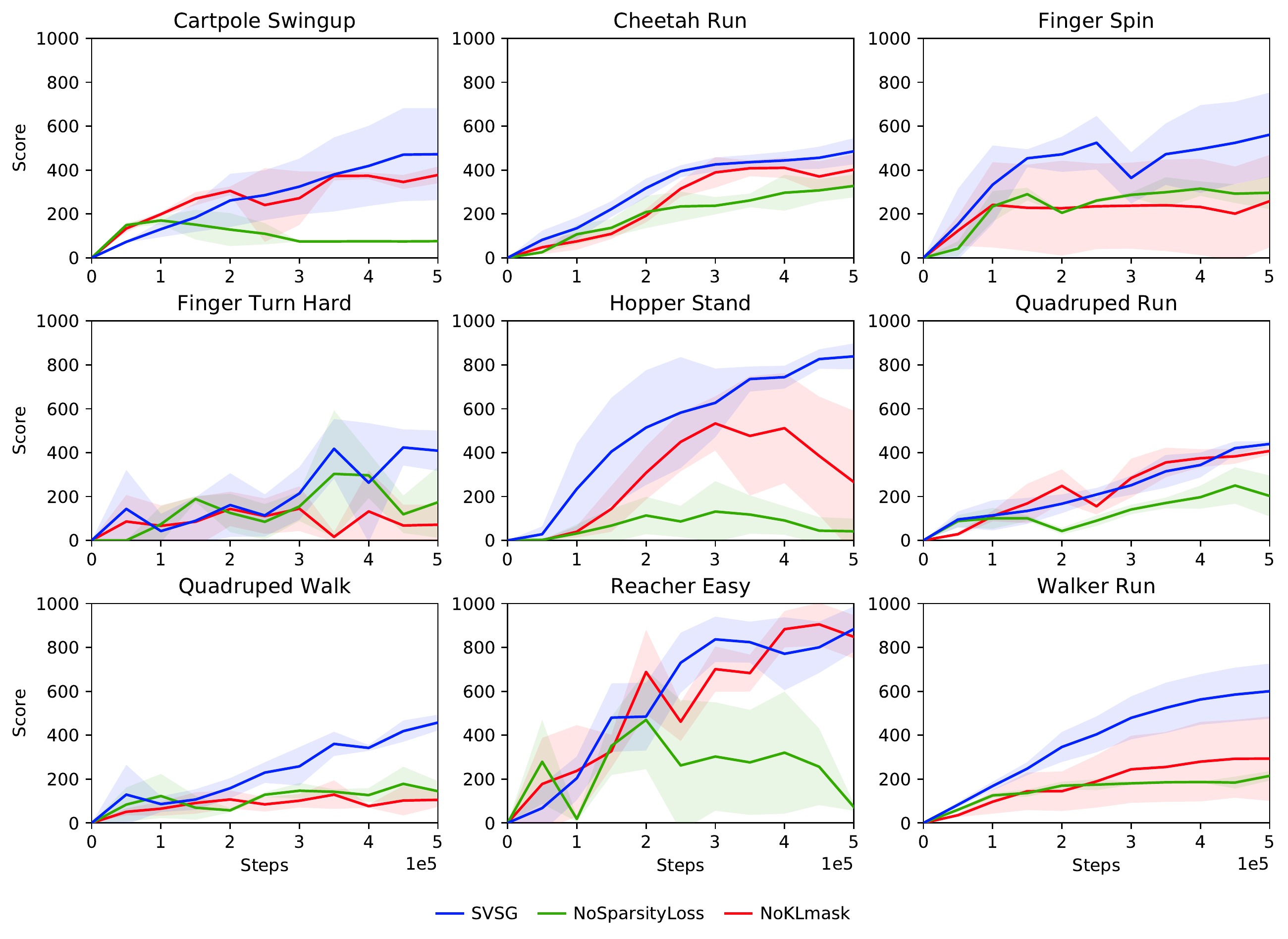}
    \caption{Ablation study comparing the performance of different SVSG models on DMC. We compare training with both KL masking and sparsity penalty (\textit{SVSG}), with only sparsity penalty (\textit{NoKLmask}), and with only KL masking (\textit{NoSparsityLoss}).}
    \label{fig:ablations_svsg}
\end{figure}
\newpage
\section{DMC with Natural Background}
\label{app:nat}
\begin{figure*}[h]
    \centering
\includegraphics[width=0.99\textwidth]{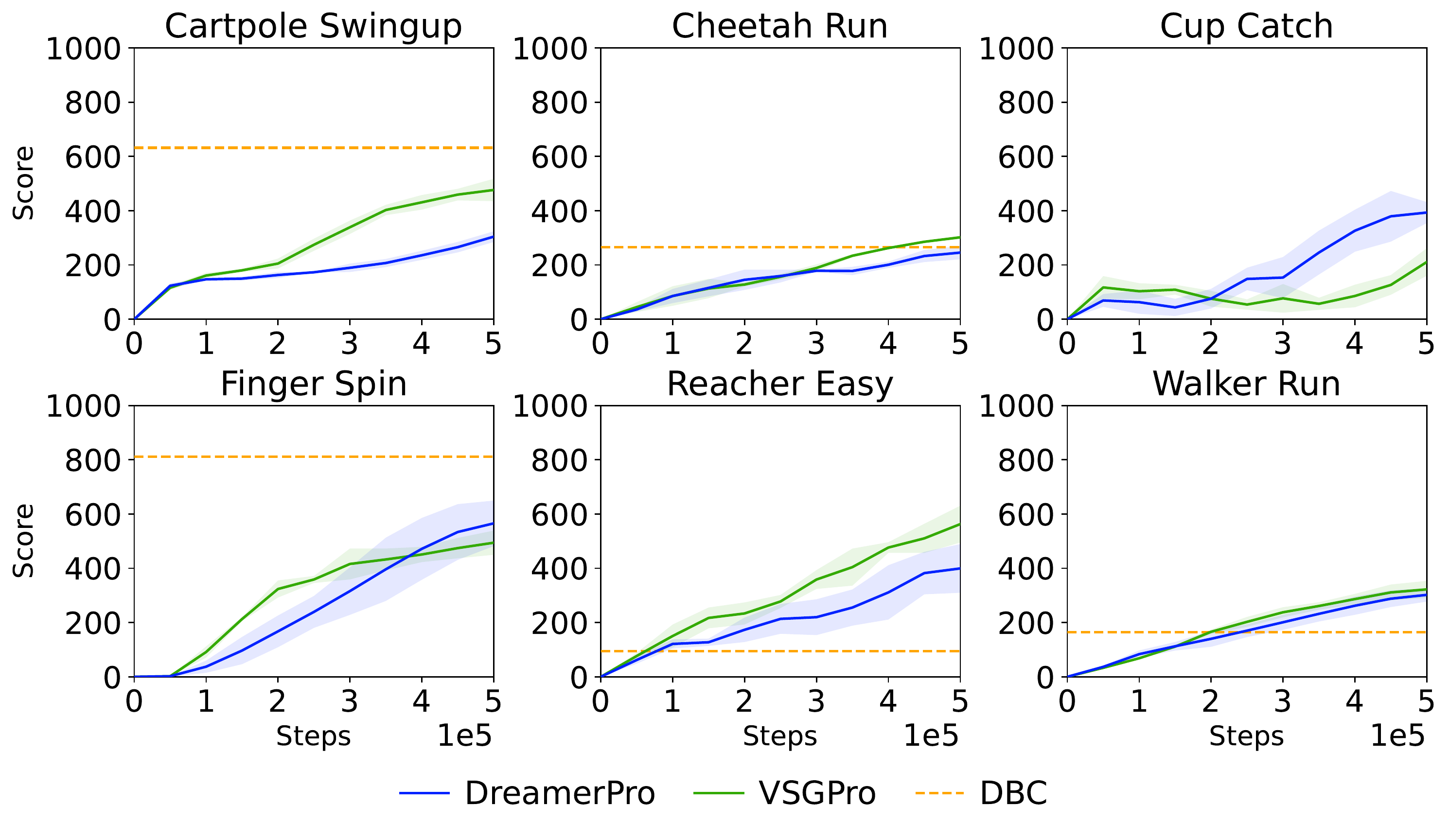}
    \caption{Results on DMC with Natural background setting.}
    \label{fig:dmc_results_plots_1M}
\end{figure*}

To evaluate the efficacy of sparse gating mechanism in  another setting with noise, we also experimented on DMC with noisy background~\cite{zhang2021learning,nguyen2021temporal}. Reconstruction free model-based RL have been found to perform better on tasks with distractive backgrounds as they don't have reconstructive loss to generate the noisy frames. We updated the RSSM in DreamerPro~\cite{deng2021dreamerpro} with \alg~and call it VSGPro. VSGPro was found to work better or similar to DreamerPro. We also compare with DBC~\cite{zhang2021learning}, which uses bisimulation metrics to learn efficient encoders that can filter noise and focus on task relevant details. Upon evaluation at 500K environment steps,  VSGPro performs better on 3 tasks and comparable on 2 tasks. Furthermore, VSGPro was found to perform similar to or better than DreamerPro. This demonstrates that the proposed gating mechanism also helps to learn efficient representations in settings with background noise. 
\newpage
\section{Atari}
\label{app:atari}
\begin{figure*}[h]
    \centering
    \includegraphics[width=\textwidth]{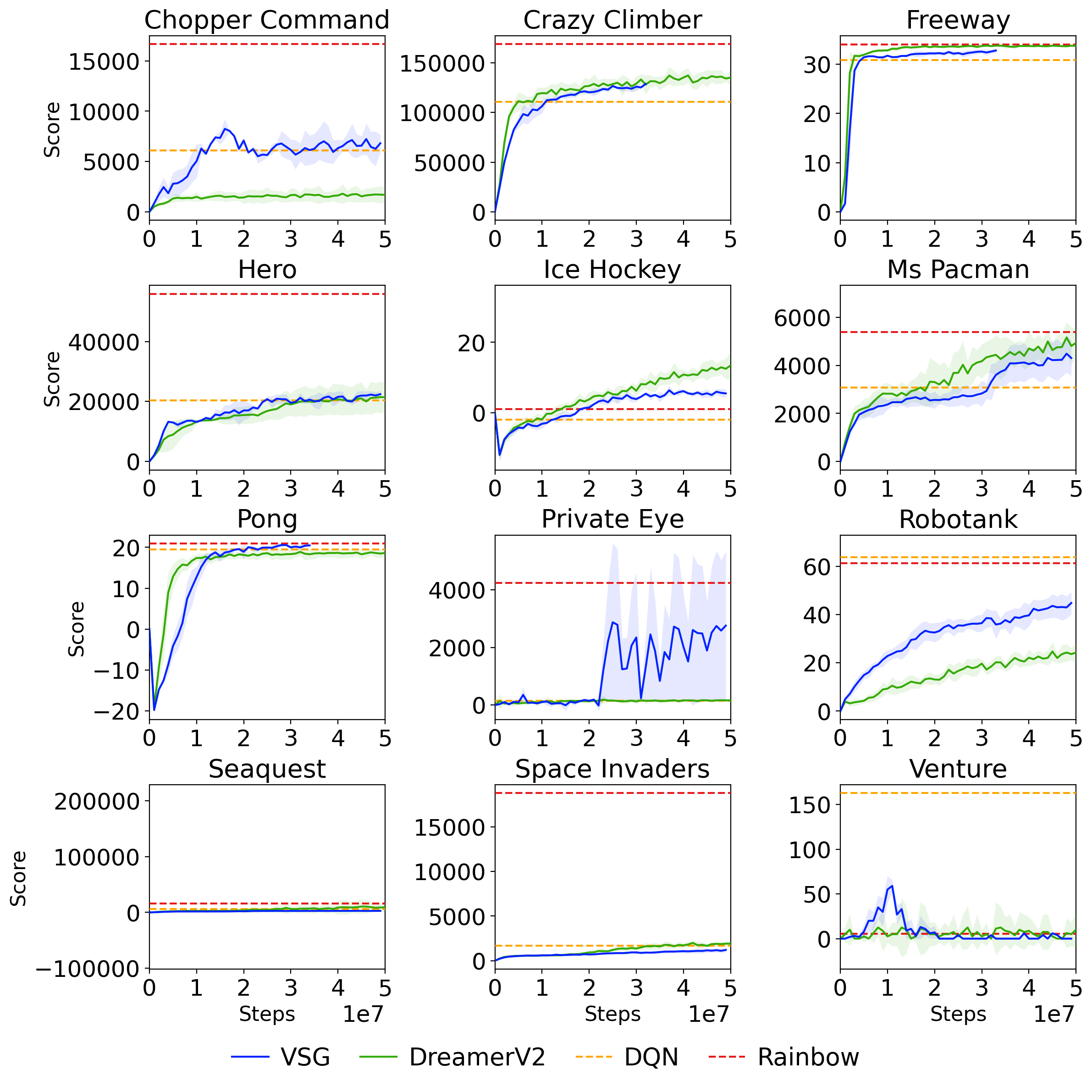}
    \caption{Results on a few tasks from the Atari benchmark trained for upto 50M environment steps.}
    \label{fig:res_atari}
\end{figure*}
In this work, we also present results on 12 tasks from the Atari benchmark~\cite{bellemare2013ale}. We trained the models for upto 50M environment steps which took around 2 days on a single NVIDIA A100 GPU for each seed. We also present results of Rainbow and DQN which were trained for 200M environment steps. For this experiment, we used the hyperparameters mentioned in the DreamerV2 paper, and use the same parameters for the gating mechanism as mentioned in Table~\ref{tab:appendix_hyperparameter}. It can be observed that on most tasks, the proposed method VSG performs similar to DreamerV2. However, we observe performance gains on Chopper Command and Robotank, and we believe that this was because in those games the viewpoint of the agent changes with movement. For example, in Robotank environment, the agent can rotate around to search for enemy tanks to shoot. VSG was performing worse than the baseline on stochastic environments- Seaquest and Ms Pacman. Also, DreamerV2 was not performing well on the Private Eye environment. However, VSG was able to learn to solve the task for a few seeds as the environment is partially-observable and agent has to enter and exit different parts of the game. We ran with 3 more seeds and a similar trend was observed where some of the seeds were failing, whereas on a few of them the model learned to solve the task.
\newpage
\section{Comparison with Stochastic State-Space Models (SSM)}
\label{app:ssm}

\begin{figure*}[h]
    \centering
    \includegraphics[width=\textwidth]{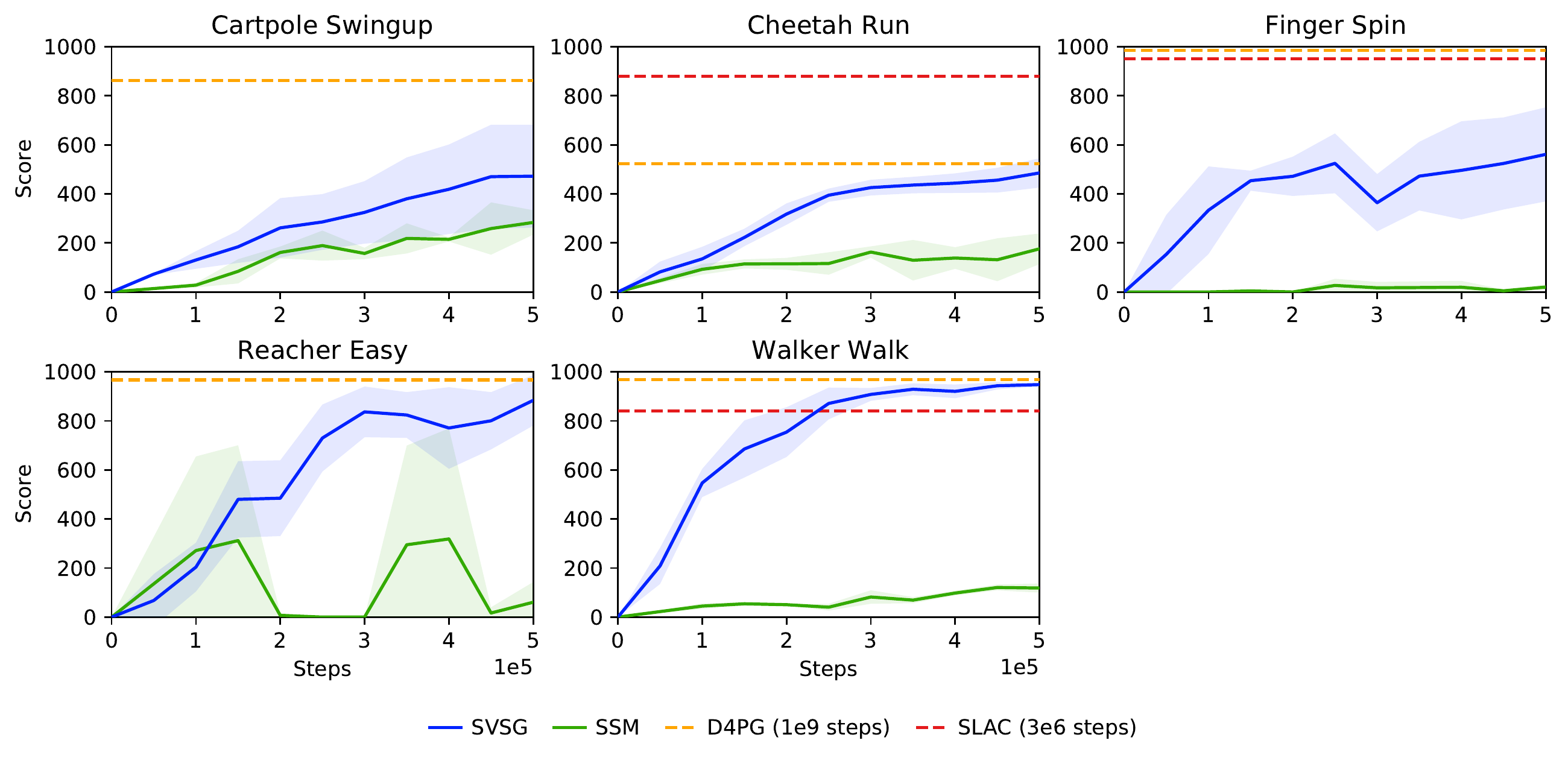}
    \caption{Comparison of SVSG and SSMs on a few tasks from the DMC trained for 500K environment steps.}
    \label{fig:res_atari}
\end{figure*}

Stochastic State-Space Models (SSMs) were discussed in PlaNet ~\cite{hafner2019planning} where the authors showed that SSMs do not achieve comparable performance when compared with RSSMs. We have shown that SVSG with a purely stochastic path can achieve comparable performance and outperform RSSMs on continuous control tasks with partial-observability and stochasticity. We also compare with SSMs as a baseline with a pure stochastic path only. In PLaNet, Cross Entropy Method~(CEM)~\cite{rubinstein1997optimization, chua2018deep} was used for planning. Since Dreamer agents improve upon PLaNet by having actor-critic network with learnable parameters in the policy and having KL-balancing in the training objective, we also implemented SSMs with those modifications. However, SSMs with those modifications were not found to work well as the actor was diverging. We also tried increasing the size of the stochastic state to larger values as it is 30 in the original implementation. We believe that sparse update prior is enabling the SVSG model to have large state sizes. Thus, we use the original implementation of SSMs from PLaNet for comparison. We experimented with a few tasks from the DMControl Suite and used 3 seeds for each task. As discussed earlier, SVSG was found to significantly outperform SSMs on all the tasks.
\newpage
\section{Comparison of Architectures}
\label{app:architectures}

\subsection{\glsfirst{rssm}}
\label{app:ablations_vsg}
\begin{figure}[hbt!]
    \centering
    \includegraphics[width=0.7\columnwidth]{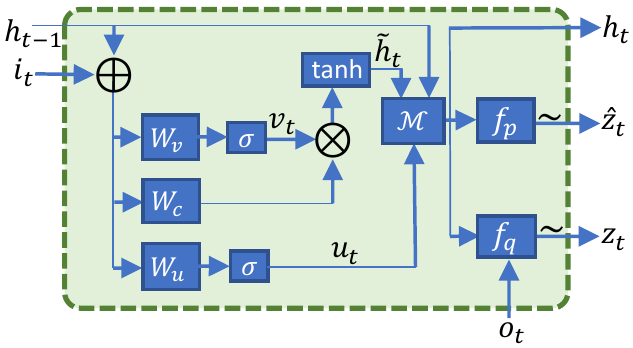}
    \caption{Architecture of \glsfirst{rssm}. $\sigma$ and $\operatorname{tanh}$ denote the sigmoid and hyperbolic tangent non-linear activation, respectively. $W_*$ and $b_*$ are the corresponding weights and biases. $\sim$, $\oplus$ and $\otimes$ denote sampling, vector concatenation, and element-wise multiplication, respectively. $\mathcal{M}$ computes $h_t=u_t\tilde{h}_t + (1-u_t) h_{t-1}$. $f_p$ and $f_q$ denote the prior and posterior distributions with learned parameters, respectively.}
    \label{fig:app_rssm}
\end{figure}

The \glsfirst{rssm} comprises of a recurrent path and an image representation path. Similar to VSG, the input~$i_t$ to the recurrent model contains information about the action and is obtained by concatenating the previous image representation state~$z_{t-1}$ and action~$a_t$ followed by passing them through through a MLP layer. \gls{rssm} uses a \glsfirst{gru}~\citep{cho2014gru} for the recurrent state where the module has two gates- reset and update.  The reset gate~$v_t$ controls the flow of information from the previous state and input, and the update gate~$u_t$ controls the extent of update of the recurrent state. Unlike VSG, the update gate in \gls{rssm} is not binary and can have values between 0 and 1. The equations are as follows:
\eq{
    v_t &= \sigma(W_v^T [h_{t-1}, i_t] + b_v)\\
    u_t &= \sigma(W_u^T [h_{t-1}, i_t] + b_u)\\
    \tilde{h}_t &= \operatorname{tanh}(v_t * (W_c^T [h_{t-1}, i_t] + b_c))\\
    h_t &= u_t \odot \tilde{h}_t + (1 - u_t) \odot h_{t-1},
}
where $\odot$ denotes element-wise multiplication, $\sigma$ and $\operatorname{tanh}$ are the sigmoid and hyperbolic tangent non-linear activation, and $W_{*}$ and $b_{*}$ denotes the weights and biases, respectively. The recurrent state is further used to obtain the posterior~$z_t$ and prior states~$\hat{z}_t$ by passing it through \gls{mlp} layers with and without observation~$o_t$, respectively.

\newpage
\subsection{\glsfirst{alg}}
We present the zoomed in architecture of \glsfirst{alg}. For details, refer to Section~\ref{sec:vsg} in the main paper.
\label{app:ablations_vsg}
\begin{figure}[hbt!]
    \centering
    \includegraphics[width=0.7\columnwidth]{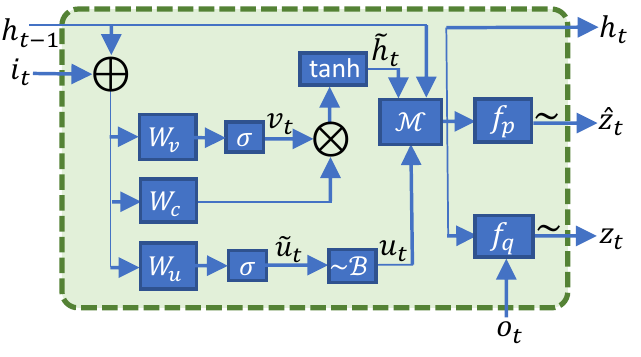}
    \caption{Architecture of \glsfirst{alg}. $\sigma$ and $\operatorname{tanh}$ denote the sigmoid and tanh non-linear activations, respectively. $W_*$ and $b_*$ are the corresponding weights and biases. $\sim$, $\oplus$ and $\otimes$ denote sampling, vector concatenation, and element-wise multiplication, respectively. $\mathcal{M}$ computes $h_t=u_t\tilde{h}_t + (1-u_t) h_{t-1}$. $\mathcal{B}$ denotes Bernoulli distribution. $f_p$ and $f_q$ denote the prior and posterior distributions with learned parameters, respectively.}
    \label{fig:app_rssm}
\end{figure}

\subsection{\glsfirst{algsimple}}
We present the zoomed in architecture of \glsfirst{algsimple}. For details, refer to Section~\ref{sec:svsg} in the main paper.
\label{app:ablations_vsg}
\begin{figure}[hbt!]
    \centering
    \includegraphics[width=0.7\columnwidth]{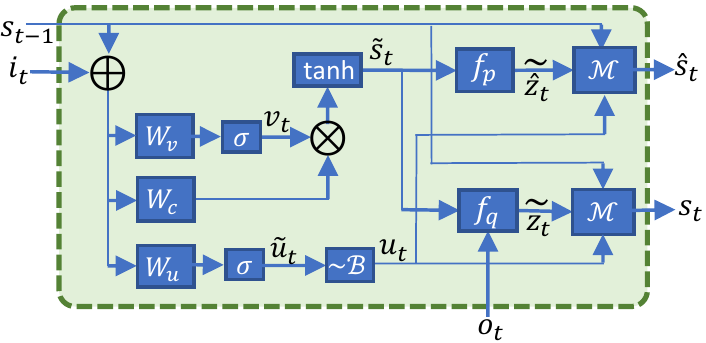}
    \caption{Architecture of \glsfirst{algsimple}. $\sigma$ and $\operatorname{tanh}$ denote the sigmoid and tanh non-linear activations, respectively. $W_*$ and $b_*$ are the corresponding weights and biases. $\sim$, $\oplus$ and $\otimes$ denote sampling, vector concatenation, and element-wise multiplication, respectively. $\mathcal{M}$ computes $s_t=u_t\tilde{s}_t + (1-u_t) s_{t-1}$. $\mathcal{B}$ denotes Bernoulli distribution. $f_p$ and $f_q$ denote the prior and posterior distributions with learned parameters, respectively.}
    \label{fig:app_rssm}
\end{figure}
\newpage
\section{Ablation for Sparse Gating Mechanism}
\label{app:latent}
To conduct this experiment, the learned world model of corresponding agents are given the first 15 frames and 5 rollouts are generated in the latent space for the next 35 frames. The sequence of actions is kept fixed across the generated trajectories, and obtained model states are passed through the decoder to reconstruct the frames. Upon observing the imagined rollouts, VSG and SVSG were demonstrated to have better memory at remembering the color and location of objects. Whereas the world model from DreamerV2 agent was found to distort the shapes and change the color of objects. This demonstrates that the proposed mechanism helps retain information for longer time steps. Please refer to supplementary material for more videos on the same.

\begin{figure*}[h]
    \centering
\includegraphics[width=0.99\textwidth]{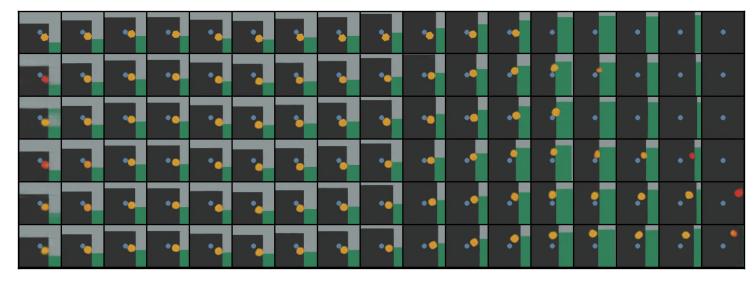}
    \caption{Imagined trajectories from the DreamerV2 agent. The top row is the ground truth and the next 5 rows are different rollouts given the same first 15 frames and action sequence. In this figure, we can observe that the model changes the color of the object towards the end of the episode.}
    \label{fig:dv2_1}
\end{figure*}

\begin{figure*}[h]
    \centering
\includegraphics[width=0.99\textwidth]{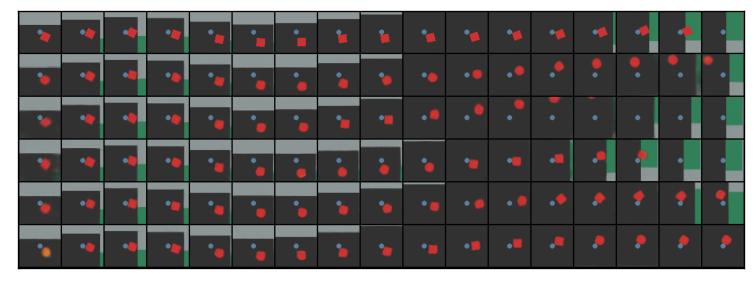}
    \caption{Imagined trajectories from the \alg~agent. The top row is the ground truth and the next 5 rows are different rollouts given the same first 15 frames and action sequence. Here, we observe that the agent is able to retain the color and shape of red block, and also reaches the final position which is close to the goal on 4 trajectories. }
    \label{fig:vsg_1}
\end{figure*}

\begin{figure*}[h]
    \centering
\includegraphics[width=0.99\textwidth]{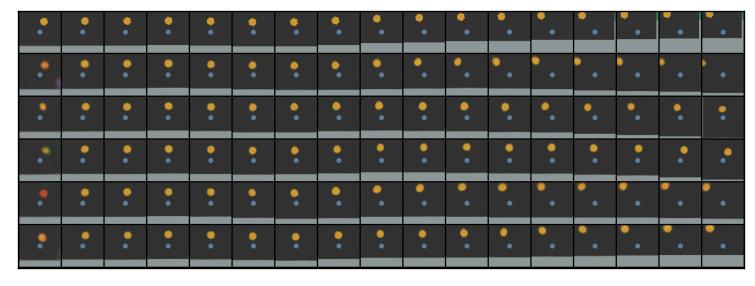}
    \caption{Imagined trajectories from the \algsimple~agent. The top row is the ground truth and the next 5 rows are different roll outs given the same first 15 frames and action sequence. We can see that SVSG is also able to retain the shape and color of objects.}
    \label{fig:svsg_1}
\end{figure*}


\end{document}